\documentclass[11pt]{article}
\pdfoutput=1

\usepackage[final]{acl} 

\usepackage{times}
\usepackage{latexsym}
\usepackage{tcolorbox}
\tcbuselibrary{skins,breakable}
\usepackage[T1]{fontenc}

\usepackage[utf8]{inputenc}

\usepackage{microtype}
\usepackage{booktabs}
\usepackage{tabularx}
\usepackage{array}
\usepackage{inconsolata}
\usepackage{subcaption}
\usepackage{graphicx}
\graphicspath{{./}{figures/}}
\usepackage{enumitem}

\usepackage{amsmath}
\usepackage{amssymb}
\usepackage{hyperref}

\title{Large Language Models in Resolving Contextual Knowledge Conflicts}

\author{\vspace{8pt}
Xinye Yang\textsuperscript{1} \hspace{2em} Zhenyang Liu\textsuperscript{1} \hspace{2em} Ruisi Li\textsuperscript{2} \hspace{2em} Yuanyuan Lei\textsuperscript{3}\\
       \textsuperscript{1}Northwestern University, Evanston, IL, \textsuperscript{2}New York University, New York, NY\\ \textsuperscript{3}Computer \& Information Science and Engineering, University of Florida, Gainesville, FL\\
       \texttt{XinyeYang2027@u.northwestern.edu}, \texttt{yuanyuan.lei@ufl.edu}}

\begin{document}
\maketitle
\begin{abstract}


Most prior works focused on conflicts between an LLM’s internal parametric knowledge and externally provided context. In contrast, we investigate how LLMs handle conflicts that arise within contextual knowledge itself. We introduce a taxonomy of six types of contextual conflicts (\textit{misinformation, inferential, temporal, granularity, perspective, and ambiguity}) and contribute a comprehensive dataset \textsc{ContextConflict} for this setting. The dataset contains 5,781 samples, covers both reasoning and summarization tasks, and includes both explicit contradictions and implicit conflicts that require multi-step reasoning. Experiments on seven LLMs show that current models still fall short in resolving contextual knowledge conflicts. We further provide mechanistic interpretability insights into how LLMs process such conflicts, revealing their latent awareness of conflicts and the representational geometry underlying conflict processing. In addition, our analysis uncovers a consistent model bias towards earlier evidence, and this positional preference serves as a key obstacle to effective conflict resolution. Motivated by these findings, we further propose a simple training-free, label-free  steering method that steers activations to encourage a more comprehensive incorporation of evidences for better conflict resolution. On our dataset, the method consistently improves accuracy on reasoning tasks and generates higher-quality, more balanced summaries for summarization tasks. \footnote{The link for dataset and code is:  \url{https://github.com/lei-nlp-lab/context_conflict_emnlp_2026}.}\footnote{\textsc{ContextConflict} dataset is also released on Hugging Face: \url{https://huggingface.co/datasets/AsherYang/ContextConflict}}

\end{abstract}

\section{Introduction}
\label{sec:intro}

\begin{figure}[t]
  \centering
  \includegraphics[width=1.0 \columnwidth]{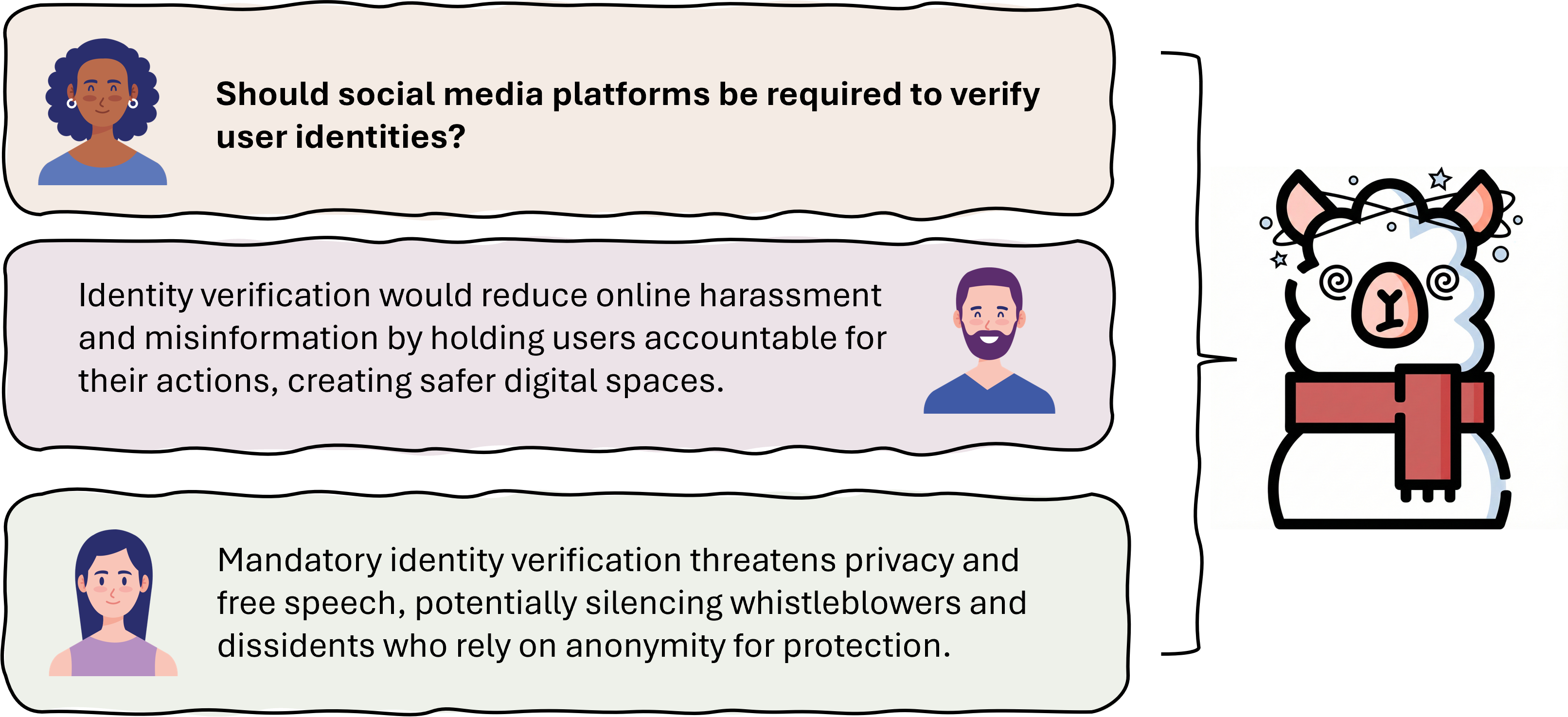}
  \caption{Example of contextual knowledge conflict: two contexts provide divergent perspectives.}
  \label{fig:knowledge_conflict_example}
\end{figure}

Large language models (LLMs) deployed in retrieval-augmented or multi-document settings routinely encounter conflicting information from multiple sources. Figure~\ref{fig:knowledge_conflict_example} illustrates a representative case: when asked about social media identity-verification policies, an LLM must synthesize opposing viewpoints that prioritize different values (safety versus privacy). The reliability of downstream applications therefore depends on how well models handle such conflicts.
When conflict resolution fails, outputs may become hallucinated, factually incorrect, or one-sided~\cite{shi-etal-2024-trusting,chen-etal-2022-rich}, with the failure mode extending to news aggregation, medical decision support, misinformation detection, and legal reasoning in high-stakes settings. Characterizing how LLMs process multi-source conflicts is therefore a prerequisite for diagnosing and correcting these failures~\cite{longpre-etal-2021-entity,chen-etal-2022-rich}.

Prior datasets on contextual knowledge conflicts exhibit four interrelated limitations. First, they rely on template-based synthetic construction, most commonly entity replacement, which fails to capture real-world complexity~\cite{NEURIPS2024_baf4b960, longpre-etal-2021-entity}. Second, they focus on explicit factual contradictions and overlook implicit conflicts that require multi-step reasoning~\cite{NEURIPS2024_baf4b960, 10.5555/3540261.3542508, Du_Bosselut_Manning_2022}. Third, they offer limited domain breadth and conflict-type coverage. Fourth, they can suffer from class imbalance, which makes category-level analysis statistically unreliable~\cite{xu-etal-2024-knowledge-conflicts, xie2024adaptive}.

To address the above gaps, we contribute \textsc{ContextConflict}, a comprehensive multi-domain dataset of contextual knowledge conflicts comprising 5,781 samples. We define a taxonomy of six conflict types, including \textit{misinformation}, \textit{inferential}, \textit{temporal}, \textit{granularity}, \textit{perspective}, \textit{ambiguity} conflicts. The first three correspond to \textbf{\textit{reasoning}} tasks, where LLMs are required to perform reasoning and identify the correct answer from multiple conflicting evidences. The latter three correspond to \textbf{\textit{summarization}} tasks, where LLMs are expected to produce balanced summaries that preserve divergent perspectives. Within each conflict type, we further distinguish \textit{explicit} conflicts that are identifiable by direct surface-level comparison, and \textit{implicit} conflicts that require multi-step cross-evidence reasoning to resolve. An evaluation of seven LLMs covering both closed-source and open-source models on this dataset show that modern LLMs still fall short in resolving contextual knowledge conflicts.



We also provide mechanistic interpretability insights into how LLMs process contextual conflicts. Specifically, we investigate the latent awareness and representational geometry of each conflict type using concept activation vectors~\cite{pmlr-v80-kim18d} and spectral energy decomposition~\cite{Eckart1936}, respectively. We also examine evidence attributions at both representation and output levels: at the representation level, we compare the activation geometry induced by single evidence versus combined evidence; at the output level, we quantify evidence contributions using Shapley-based attribution scores~\cite{lundberg2017unifiedapproachinterpretingmodel}. Our analysis reveals strong awareness of contextual conflicts internalized in models, with different conflict types emerging at different layer depths and represented as distinct geometry in model's latent space. We also uncover a consistent model bias favoring earlier-positioned evidences, suggesting that this positional preference is a key obstacle to comprehensive evidence integration.

To address this issue, we then design a training-free, label-free steering method that mitigates positional preference and encourages more comprehensive consideration of conflicting evidences. Specifically, we construct a steering direction by computing the centroid of LLMs activations when processing each piece of evidence individually. During inference, we nudge the model’s activations along this steering direction, guiding it to attend more evenly across evidence positions and thereby integrate evidence more comprehensively. The results show that our simple method effectively improves LLMs’ ability to resolve knowledge conflicts, yielding more balanced summaries in summarization tasks and higher accuracy on reasoning tasks.


Our main contributions are summarized below.
\begin{itemize}
    \item We contribute a comprehensive multi-domain dataset covering six types of contextual knowledge conflicts, spanning reasoning and summarization tasks across diverse domains
    \item We present a mechanistic interpretability analysis of contextual conflict processing, revealing how each conflict are detected, geometrically represented, and processed within LLMs
    \item We design a training-free, label-free steering method that encourages more comprehensive integration of conflicting evidences, yielding better performance in conflict resolution
\end{itemize}

\section{\textsc{ContextConflict}: Contextual Knowledge Conflict Dataset}

\subsection{Overview}
\label{sec:dataset_overview}

We contribute a multi-domain contextual knowledge conflict dataset to address the limitations of existing datasets. We define a taxonomy of six conflict types and each conflict type probes a distinct capability, including \textit{inferential} reasoning (multi-step reasoning over conflicting premises), \textit{misinformation} robustness (resistance to plausible disinformation), \textit{temporal} reasoning (tracking claim validity over time), \textit{granularity} alignment (reconciling information at different abstraction levels), \textit{perspective} integration (balancing diverse viewpoints), and \textit{ambiguity} resolution (disambiguating co-referential entities across documents). These six conflict types can be organized into two task families: \textbf{Reasoning} tasks (\textit{inferential}, \textit{misinformation}, \textit{temporal}) require models to identify the correct answer under conflicting evidence, and \textbf{Summarization} tasks (\textit{ambiguity}, \textit{perspective}, \textit{granularity}) require models to produce balanced summaries that preserve divergent viewpoints. Within each family, conflicts are further divided into \textbf{explicit} and \textbf{implicit} cases. Explicit cases are detectable by direct surface-level comparison and require at most one inferential step. Implicit cases emerge through multi-step cross-evidence reasoning and require at least two inferential steps. We provide the per-type implicit proportions in Table~\ref{tab:dataset_composition}.


The desired model output differs by task family. In Reasoning tasks, each instance has a verifiable ground-truth answer, so the model is expected to identify the correct conclusion under the conflicting evidence, and Accuracy serve as evaluation metric. In Summarization tasks, each instance admit multiple valid responses: \textit{perspective} and \textit{ambiguity} often involve diverging political opinions or social questions, while \textit{granularity} allows compatible answers at different specificity levels. The model is expected to comprehensively integrate the divergent viewpoints and generate a balanced summary: for \textit{perspective} conflicts, it should present and contrast all stances without privileging any; for \textit{ambiguity} conflicts, it should identify the underlying name collision and cover each referenced entity; for \textit{granularity} conflicts, it should integrate the different levels of specificity and state their compatibility. The Shapley-based Balance score is employed as evaluation metric to measure evenness of evidence use. More evaluation details are in \ref{sec:evaluation_metrics}.

\begin{table}[t]
\centering
\scriptsize
\setlength{\tabcolsep}{2pt}
\renewcommand{\arraystretch}{1.12}
\begin{tabularx}{\columnwidth}{l r c >{\centering\arraybackslash}p{1.2cm} X}
\toprule
\textbf{Type} & \textbf{Size} & \textbf{Avg.\#Ev.} & \textbf{Implicit (\%)} & \textbf{Base source(s)} \\
\midrule
Misinformation & 1004 & 2.66 & 37.1 & SciFact; ConflictBank \\
Inferential    &  787 & 5.66 & 25.4 & ENTAILMENTBANK; NEJM-MedQA; FOLIO \\
Temporal       &  960 & 2.26 & 93.5 & ConflictBank; CONFLICTS \\
Granularity    & 1020 & 5.36 &  7.4 & ROAST-ABSA; NEJM-MedQA \\
Perspective    & 1010 & 4.48 & 49.5 & AllSides; Perspectrum \\
Ambiguity      & 1000 & 2.94 &  0.0 & AmbigDocs \\
\midrule
\textbf{Total} & \textbf{5781} & & & \\
\bottomrule
\end{tabularx}
\caption{\textsc{ContextConflict} statistics: size, average number of evidence pieces, implicit conflict proportion, and base sources for each conflict type.}
\label{tab:dataset_composition}
\end{table}

\subsection{Dataset Composition}
\label{sec:construction}

We construct the dataset from ten base datasets spanning diverse domains; Appendix Table~\ref{tab:base_datasets} lists each source with its domain, license, and per-conflict-type usage. The dataset consists of 1,734 semi-synthetic instances (29.99\%) and 4,047 preserved-original instances manually categorized by conflict type (70.01\%), for a total of 5,781.

Unlike prior synthetic conflict datasets that build conflicts via entity replacement~\cite{NEURIPS2024_baf4b960, longpre-etal-2021-entity}, which keeps the surrounding context unchanged, swaps only the entity, and produces shallow lexical contradictions, our semi-synthetic instances use GPT-5~\cite{singh2025openaigpt5card} only as an auxiliary generator for conflicting evidence or timestamps; gold labels are inherited from the source dataset or human re-annotated, never produced by GPT-5. This preserves discourse-level coherence in the conflicting evidence while keeping label integrity independent of the generator.

Quality assurance is risk-proportional, scaling with how much each construction step can corrupt labels. Subsets whose perturbations may flip the gold label receive full re-annotation by two independent annotators; subsets whose augmentations rarely change the label are audited on a 30\% sample; preserved-original subsets are spot-checked at 10\%. Per-source provenance counts are released in the metadata; Appendix~\ref{sec:dataset_construction} reports the full GPT-5 prompts and validation rules, and Appendix~\ref{app:data_samples} provides representative samples.

\paragraph{Granularity and Inferential Conflicts.}
NEJM-MedQA~\cite{Savage2024} combines U.S.\ medical-licensing exam questions with real clinical cases from the \emph{New England Journal of Medicine}. Each instance provides clinical evidence (symptoms, lab results, patient history) and several candidate reasoning chains. Each chain is generated under a distinct diagnostic-reasoning prompt strategy from the source dataset and carries a binary gold-correctness flag adjudicated by clinicians. We select cases where two chains disagree on the final diagnosis. When both diagnoses are gold-correct but at different abstraction levels (e.g., a broad syndrome label versus a specific pathogen it subsumes), we label a \textit{granularity} conflict: the two answers are compatible and differ only in diagnostic specificity. When only one diagnosis is gold-correct and the other reaches a wrong conclusion through a flawed intermediate step (e.g., a misapplied clinical heuristic), we label an \textit{inferential} conflict: the disagreement is at the reasoning-process level. The original question and reasoning chains are preserved as evidence.

\paragraph{Inferential Conflicts.}
FOLIO~\cite{han-etal-2024-folio} and ENTAILMENTBANK~\cite{dalvi-etal-2021-explaining} are logical inference datasets. Each instance gives a set of premises and a hypothesis, labeled by whether the premises entail, contradict, or are neutral to the hypothesis. We add one or two extra statements that perturb the original reasoning chain. Because the added evidence can change label validity, every resulting instance is re-annotated by two independent annotators, with disagreements resolved by discussion (100\% double-annotation coverage; annotation interface in Appendix Figure~\ref{fig:folio_ann_ui}).

\paragraph{Misinformation Conflicts.}
SciFact~\cite{wadden-etal-2020-fact} is a fact-verification dataset. Each instance pairs a claim with evidence sentences from research abstracts, labeled as supporting or refuting the claim. We generate conflicting evidence in a matching writing style, including experimental-style citations, to build \textit{misinformation} conflicts. Because this augmentation rarely changes gold labels, we audit a 30\% sample for stylistic and argumentative consistency.

\paragraph{Temporal Conflicts.}
ConflictBank-temporal~\cite{NEURIPS2024_baf4b960} instance contains evidence statements with an implicit chronological order and a time-sensitive question. We attach an explicit timestamp to each statement, without modifying the original evidence, so that temporal relationships become unambiguous. Because timestamps can change gold labels, each instance is verified by two independent annotators with discussion-based adjudication.

\paragraph{Curation of Remaining Datasets.}
For datasets used without synthetic generation (AmbigDocs, ROAST-ABSA, AllSides, Perspectrum, CONFLICTS), we randomly sample 10\% of each source for quality review and conflict-category validation.

\section{Evaluation}

\subsection{Evaluation Metrics}
\label{sec:evaluation_metrics}

We evaluate model performance using complementary metrics with task-specific emphasis.

\noindent\textbf{Accuracy.}
For \textbf{reasoning tasks} (\textit{inferential}, \textit{misinformation}, and \textit{temporal} conflicts), we measure accuracy as the proportion of model outputs that match the gold-standard labels.

\noindent\textbf{Evidence Balance.}
For \textbf{summarization tasks}, all evidence sources are equally valid despite their conflicting content, so we quantify how evenly a response integrates them with a Shapley-based attribution framework. We assign each evidence piece a Shapley-value contribution to the response's likelihood and then measure how unequally these contributions are distributed.
Given $n$ evidence pieces, index set $N=\{1,\ldots,n\}$, and model response $R$, the marginal contribution of piece $i$ is
\begin{equation}
\phi_i = \sum_{S \subseteq N \setminus \{i\}} \frac{1}{n\binom{n-1}{|S|}} \bigl( v(S \cup \{i\}) - v(S) \bigr),
\end{equation}
where $v(S) = \frac{1}{|R|}\sum_{t=1}^{|R|}\log p_{\text{scorer}}(r_t \mid r_{<t}, \mathcal{E}_S)$ is the length-normalized log-likelihood of $R$ under a frozen external scorer (default: Llama-3.2-1B; scorer-size sensitivity is examined in Section~\ref{sec:balance_validation}).
We clip negative contributions, which arise when the response contradicts a piece, and normalize the non-negative mass into a share distribution $p_i = \max(0,\phi_i) / \sum_{j}\max(0,\phi_j)$. We then report Balance as the normalized Gini coefficient of $\mathbf{p}$:
\begin{equation}
\text{Balance}(R)=\frac{n}{n-1}\cdot\frac{1}{n}\sum_{i=1}^{n}(2i - n - 1)\,p_{[i]},
\end{equation}
where $p_{[1]} \leq \cdots \leq p_{[n]}$ are sorted contributions. Lower scores indicate more balanced integration (0 = perfect equality; 1 = maximum inequality).

\noindent\textbf{Faithfulness.}\footnote{Computed using RAGAS~\cite{es-etal-2024-ragas}: \url{https://github.com/vibrantlabsai/ragas}} To detect hallucinated content that goes beyond the given evidence in conflict scenarios, we report Faithfulness, applicable to all conflict types. It decomposes a response into atomic claims and reports the proportion judged supported by the given evidence. Faithfulness measures grounding rather than factual correctness: a factually accurate response still scores low if some claims are not grounded in the provided evidence.

\subsection{Validating the Balance Metric}
\label{sec:balance_validation}

Unlike established metrics such as Accuracy and Faithfulness, Evidence Balance relies on our Shapley-based attribution framework. We confirm its reliability with three independent checks: a human and LLM agreement study, a causal intervention on the same human-verified subset, and a scorer-size robustness check.

\noindent\textbf{Human Evaluation and LLM-as-a-Judge.}
To assess agreement between our automatic metric and human judgment of evidence balance, we collect annotations from two independent humans and a GPT-5 LLM-as-a-Judge on 108 samples spanning three models and three summarization tasks. Our metric reaches $\kappa=0.50$--$0.52$ against the human annotators (Table~\ref{tab:kappa_agreement}), comparable to inter-human agreement ($\kappa=0.52$) and to the LLM-as-a-Judge ($\kappa=0.51$). Annotation setup, interface, and per-model Balance appear in Appendix~\ref{human_eval}.

\noindent\textbf{Causal Intervention.}
To probe whether Shapley rankings reflect actual evidence dependence, we run a causal-intervention check on the human-annotated subset above, restricted to the two open-weight models where each model's own log-probabilities are accessible (Llama-3.1-8B-Instruct and GPT-OSS-20B), giving 72 already human-verified samples. We remove the highest- and lowest-contributing evidence piece and measure the change in each model's own log-likelihood. As Table~\ref{tab:intervention} shows, removing the highest-contributing evidence causes substantially larger drops than removing the lowest, indicating that attribution rankings align with actual evidence dependence.

\noindent\textbf{Scorer-Size Robustness.}
To rule out artifacts of the 1B default scorer, we rerun the Shapley computation with two alternative scorers on the three summarization tasks: one larger from the same family (Llama-3.1-8B) and one comparable in size from a different family (Gemma-2B). As shown in Appendix Table~\ref{tab:bal_robustness_two_scorers}, absolute Balance values shift but the ranking of evaluated models is essentially unchanged, confirming that our results do not depend on the default scorer.

\begin{table}[t]
\centering
\footnotesize
\begin{tabular}{lc}
\toprule
\textbf{Annotator Pair} & \textbf{Cohen's Kappa} \\
\midrule
\multicolumn{2}{l}{\textit{Human $\leftrightarrow$ Human}} \\
Human 1 vs Human 2          & 0.52 \\
\addlinespace[2pt]
\multicolumn{2}{l}{\textit{Human $\leftrightarrow$ Ours}} \\
Human 1 vs Ours        & 0.52 \\
Human 2 vs Ours        & 0.50 \\
\addlinespace[2pt]
\multicolumn{2}{l}{\textit{Human $\leftrightarrow$ LLM-as-a-Judge}} \\
Human 1 vs LLM-as-a-Judge   & 0.55 \\
Human 2 vs LLM-as-a-Judge   & 0.56 \\
\addlinespace[2pt]
\multicolumn{2}{l}{\textit{Ours $\leftrightarrow$ LLM-as-a-Judge}} \\
LLM-as-a-Judge vs Ours & 0.51 \\
\bottomrule
\end{tabular}
\caption{Inter-annotator agreement (Cohen's Kappa).}
\label{tab:kappa_agreement}
\end{table}

\begin{table}[t]
\centering
\scriptsize
\setlength{\tabcolsep}{3pt}
\begin{tabular}{lccccc}
\toprule
\textbf{Model} & $\ell_{\text{full}}$ & $\ell_{\text{high}}$ & $\ell_{\text{low}}$ & $\Delta_{\text{high}}$ & $\Delta_{\text{low}}$ \\
\midrule
Llama-3.1-8B & $-0.656$ & $-0.977$ & $-0.732$ & $0.321$ & $0.076$ \\
GPT-OSS-20B  & $-6.557$ & $-7.442$ & $-6.930$ & $0.885$ & $0.373$ \\
\bottomrule
\end{tabular}
\caption{Causal intervention on the 72 human-verified samples. $\Delta_{\text{high}} {=} \ell_{\text{full}} {-} \ell_{\text{high}}$, $\Delta_{\text{low}} {=} \ell_{\text{full}} {-} \ell_{\text{low}}$, using each model's own log-likelihood.}
\label{tab:intervention}
\end{table}

\subsection{Evaluation Results}
\label{eval_result}
\begin{table*}[t]
\centering
\scriptsize
\setlength{\tabcolsep}{2.5pt}
\renewcommand{\arraystretch}{1.0}
\newcolumntype{Y}{>{\centering\arraybackslash}X}

\begin{tabularx}{\textwidth}{l *{12}{Y}}
\toprule
& \multicolumn{6}{c}{\textbf{Reasoning-based Tasks}} & \multicolumn{6}{c}{\textbf{Summarization-based Tasks}} \\
\cmidrule(lr){2-7}\cmidrule(lr){8-13}
\textbf{Model} &
\multicolumn{2}{c}{\textbf{Infer.}} &
\multicolumn{2}{c}{\textbf{Misinfo.}} &
\multicolumn{2}{c}{\textbf{Temp.}} &
\multicolumn{2}{c}{\textbf{Ambig.}} &
\multicolumn{2}{c}{\textbf{Gran.}} &
\multicolumn{2}{c}{\textbf{Persp.}} \\
\cmidrule(lr){2-3}\cmidrule(lr){4-5}\cmidrule(lr){6-7}\cmidrule(lr){8-9}\cmidrule(lr){10-11}\cmidrule(lr){12-13}
& Acc$\blacktriangle$ & Fth$\blacktriangle$
& Acc$\blacktriangle$ & Fth$\blacktriangle$
& Acc$\blacktriangle$ & Fth$\blacktriangle$
& Bal$\blacktriangledown$ & Fth$\blacktriangle$
& Bal$\blacktriangledown$ & Fth$\blacktriangle$
& Bal$\blacktriangledown$ & Fth$\blacktriangle$ \\
\midrule
gpt-5             & 43.6 & 76.8 & \textbf{61.5} & 83.2 & \textbf{68.0} & \underline{88.5} & 26.1 & 81.8 & \textbf{20.2} & \textbf{91.1} & \textbf{29.8} & 71.5 \\
claude-4.5-sonnet & \textbf{44.3} & 72.8 & \underline{54.6} & 81.7 & 61.9 & \textbf{90.7} & \underline{24.3} & \underline{83.8} & 22.9 & 87.5 & \underline{31.0} & 63.7 \\
gemini-2.5-pro    & 42.4 & \textbf{80.0} & 38.6 & \textbf{89.9} & \underline{65.7} & 87.1 & \textbf{20.5} & \textbf{92.2} & 25.9 & 85.4 & 33.9 & 70.4 \\
gpt-oss-120b      & \underline{44.1} & 74.2 & 51.4 & \underline{84.4} & 64.5 & 82.8 & 31.9 & 74.4 & 23.8 & 81.6 & \underline{31.0} & \underline{79.7} \\
gpt-oss-20b       & 35.7 & 72.3 & 51.8 & 82.8 & 63.2 & 84.5 & 39.8 & 75.3 & \underline{21.8} & \underline{88.5} & 34.5 & \textbf{92.5} \\
llama-3.1-70b-instruct     & 38.4 & \underline{79.5} & 46.0 & 80.3 & 49.7 & 84.6 & 27.7 & 81.2 & 30.3 & 77.1 & 32.4 & 79.4 \\
llama-3.1-8b-instruct      & 24.8 & 61.6 & 27.5 & 68.2 & 19.5 & 53.7 & 41.5 & 78.5 & 34.4 & 73.3 & 36.9 & 72.3 \\
\addlinespace[2pt]
\midrule
\addlinespace[2pt]
\multicolumn{13}{l}{\textit{Llama-3.1-8B-Instruct}} \\
\quad + CAS                  & \underline{34.2} & 58.7 & 31.8 & \underline{66.1} & \underline{34.7} & \underline{49.8} & 35.7 & \textbf{91.4} & 28.7 & 84.3 & 33.5 & 80.8 \\
\quad + All-Tokens           & \textbf{37.0} & \underline{59.4} & \textbf{35.4} & \textbf{66.9} & \textbf{38.7} & 49.0 & \textbf{30.8} & 89.3 & \underline{25.8} & \underline{86.1} & \textbf{27.7} & \textbf{82.6} \\
\quad + First-Generated      & 32.0 & \textbf{63.2} & \underline{32.4} & 65.0 & 25.7 & \textbf{51.2} & \underline{31.4} & \underline{89.7} & \textbf{25.7} & \textbf{88.6} & \underline{29.0} & \underline{81.8} \\
\addlinespace[2pt]
\multicolumn{13}{l}{\textit{GPT-OSS-20B}} \\
\quad + CAS                  & 42.1 & \underline{68.9} & \underline{52.6} & \underline{84.9} & \textbf{65.5} & \underline{83.5} & 33.4 & \underline{84.9} & \textbf{23.9} & \underline{85.5} & \underline{31.3} & \underline{85.6} \\
\quad + All-Tokens           & \underline{45.7} & 68.5 & \textbf{56.0} & 81.4 & \underline{64.1} & 83.2 & \underline{28.8} & 84.3 & \underline{24.0} & 83.0 & 32.0 & \textbf{88.6} \\
\quad + First-Generated      & \textbf{46.8} & \textbf{69.4} & 51.9 & \textbf{85.5} & 59.8 & \textbf{85.3} & \textbf{28.4} & \textbf{86.1} & 24.8 & \textbf{87.7} & \textbf{30.9} & 83.5 \\
\bottomrule
\end{tabularx}
\caption{Performance across all conflict tasks (\%). Acc $\blacktriangle$ and Fth $\blacktriangle$ (higher is better) are reported for reasoning; Bal $\blacktriangledown$ (lower is better) and Fth $\blacktriangle$ for summarization. All-Tokens and First-Generated are our two injection schedules of $u^{(l)}$ (\S\ref{sec:method}); CAS is the Context-Aware Steering alternative direction we compare against (\S\ref{sec:method}). Best and second-best within each model block are \textbf{bold} / \underline{underlined}.}
\label{tab:all_tasks}
\end{table*}

\textbf{Task Difficulty and Model Scaling.} Table~\ref{tab:all_tasks} reports performance across seven base models and six conflict categories.
On reasoning tasks, we observe a difficulty hierarchy: Temporal reaches the highest accuracy (up to 68.0\%), followed by Misinformation (up to 61.5\%) and Inferential (up to 44.3\%), indicating increasing difficulty. Performance scales with model capacity (Inferential: 24.8\% on llama-3.1-8b-instruct vs.\ 44.3\% on claude-4.5-sonnet), yet Inferential stays below 50\% even for the strongest model, suggesting multi-step reasoning under conflict is a shared SOTA bottleneck.

\textbf{Pervasive Position Bias.} Our Shapley-based attribution reveals strong evidence-position bias in summarization tasks. As shown in Appendix Figure~\ref{fig:evidence_order_bias_pie_chart} for Llama-3.1-8B-Instruct, Ambiguity shows the strongest bias, with the first evidence contributing 69.0\%, while Perspective and Granularity are less extreme but still imbalanced. Even top-performing models remain far from perfectly balanced integration. Detailed per-model evidence distributions are reported in Appendix~\ref{sec:evidence_order_bias_appendix}. Balance is also not monotonic in capacity (GPT-OSS-120B 31.9 vs.\ 20B 39.8), suggesting that scale alone cannot eliminate position bias and motivating the correction in Section~\ref{sec:method}.

\textbf{Faithfulness Patterns.} Faithfulness is higher on summarization than on reasoning and is not tightly coupled with Balance (Table~\ref{tab:all_tasks}): GPT-OSS-20B on Perspective reaches 92.5\% Faithfulness while its Balance stays at 34.5, showing that a model can be simultaneously well-grounded in the provided evidence and positionally biased, so the bias operates primarily at the evidence-selection stage.

\section{Analysis}
\label{sec:analysis}

We conduct three complementary analyses to understand how LLMs internally process conflicting knowledge. (i) \textbf{Conflict awareness} measures the initial detection point via the linear separability of hidden states. (ii) \textbf{Representational geometry} examines structural segregation via spectral energy analysis. (iii) \textbf{Position bias} traces how evidence order skews latent representations and final outputs via Shapley attribution. Together, these lenses provide a layered map of how conflict is encoded, organized, and resolved.

Our mechanistic analyses focus on Llama-3.1-8B-Instruct. To ensure our findings reflect fundamental cognitive mechanisms rather than architectural quirks, we replicate key trends on GPT-OSS-20B~\cite{openai2025gptoss120bgptoss20bmodel}, which differs in both lineage and scale. Appendix~\ref{sec:sea_other_models} reports the layer-wise AUC, $\Delta\text{ER}$ curves, and projections for this model. The consistent conflict-type ordering and emergence patterns across both architectures confirm that these conflict-processing mechanisms are robust, inherent behaviors of modern LLMs.

\subsection{Conflict Awareness via Concept Vectors}

\begin{figure}[ht!]
  \centering
  \includegraphics[width=0.9\columnwidth]{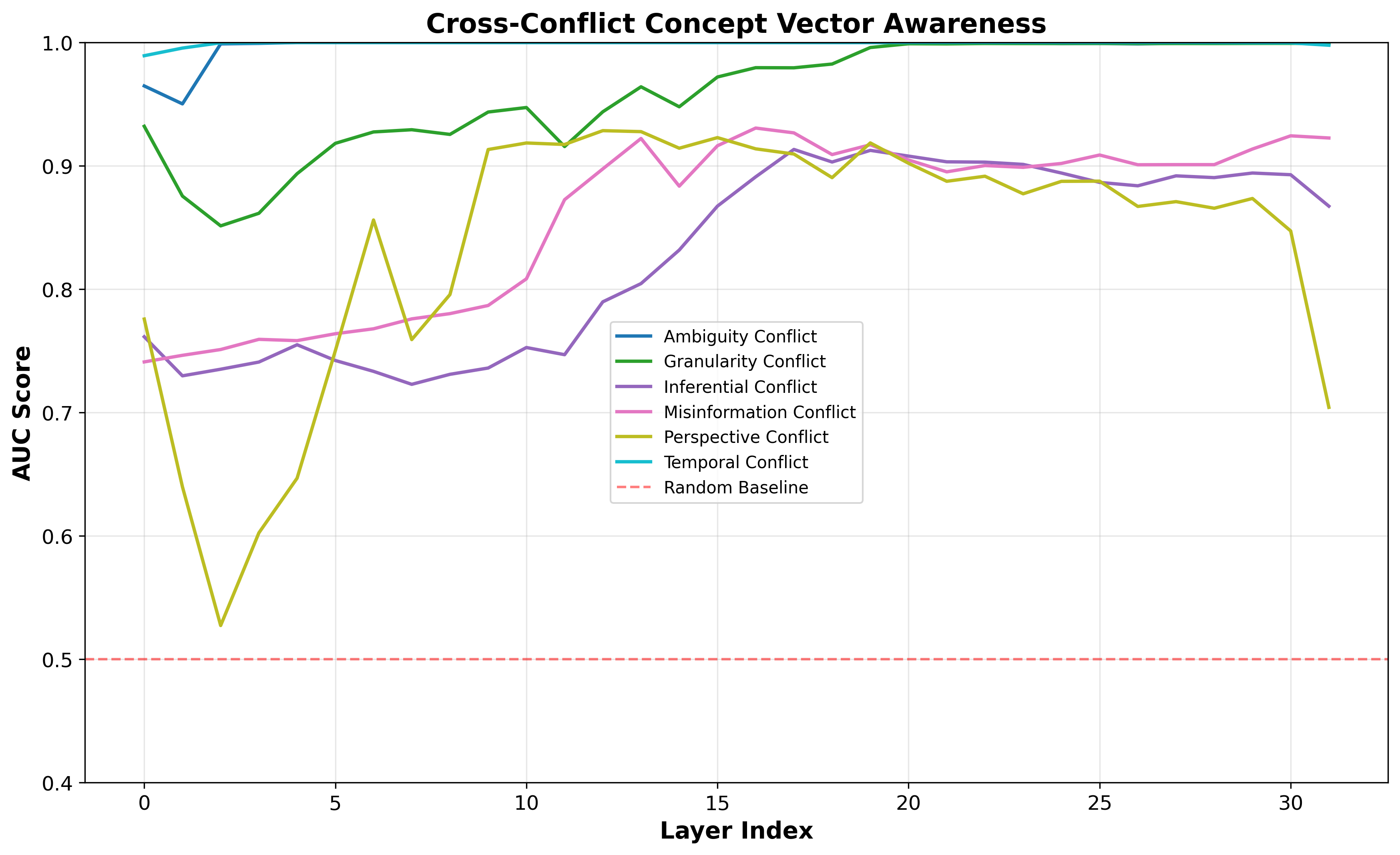}
  \caption{Layer-wise AUC for conflict vs. consistent sample classification. Higher AUC indicates stronger conflict awareness.}
  \label{fig:auc_comparison_llama8b}
\end{figure}

To understand how LLMs process confliction, we must first verify if they internally register it. Our core premise is that if a model possesses latent conflict awareness, its hidden states for conflicting inputs should be linearly separable from consistent ones. We quantify this by constructing paired datasets for each conflict type $t$, consisting of a conflict version $x_{\text{conf}}^{(t)}$ and a consistent version $x_{\text{cons}}^{(t)}$ (details in Appendix~\ref{sec:consistent_pairs}). At each layer $l$, we train a linear logistic-regression probe on $h_l(x)$ and measure the classification performance via AUC:
\begin{equation}
\text{AUC}_l^{(t)} = \text{AUC}\big(\{(h_l(x_{\text{conf}}^{(t)}), 1)\}, \{(h_l(x_{\text{cons}}^{(t)}), 0)\}\big)
\end{equation}
This approach allows us to map the precise trajectory of conflict awareness. We reveal not only if the model detects a contradiction, but exactly where and how it emerges within the architecture.

Our findings show that models exhibit robust, layer-wise conflict awareness, with signal emergence tied to semantic complexity. Figure~\ref{fig:auc_comparison_llama8b} illustrates that most conflict types reach high separability (AUC > 0.85) in mid-to-late layers, yet their developmental paths differ. \textit{Temporal} and \textit{ambiguity} conflicts saturate earliest, as they rely on explicit markers handled by early syntactic processing~\cite{tenney-etal-2019-bert}. Conversely, \textit{inferential} and \textit{misinformation} conflicts emerge later. These require multi-step reasoning or world knowledge, which rely on abstract semantic representations from deeper layers. Other types show non-monotonic patterns: \textit{granularity} follows a U-shape, while \textit{perspective} conflicts peak in middle layers before declining during viewpoint reconciliation. These trends demonstrate that conflict awareness is not a single trigger. It is a dynamic, heterogeneous process that aligns with the model's progressive semantic refinement, as confirmed across multiple models in Appendix~\ref{sec:cv_additional_models_implicit}.

\begin{figure*}[t]
  \centering
  \includegraphics[width=0.9\textwidth]{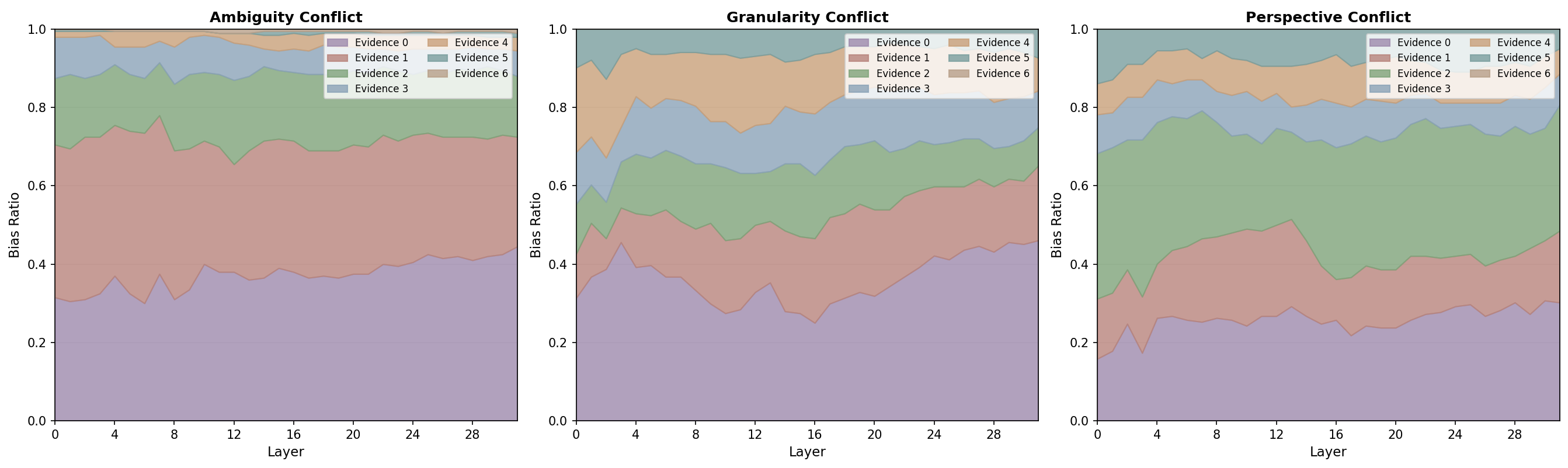}
  \caption{Layer-wise directional projection. $c^{(l)}$ consistently aligns with specific evidence directions, deviating from uniform.}
  \label{fig:bias_simple_prompt}
\end{figure*}

\subsection{Spectral Energy Analysis Reveals Conflict Dimensionality}
\label{sec:spectral_energy}

Spectral energy analysis allows us to look beyond \textit{whether} a model detects a conflict, and instead uncover \textit{how} that information is geometrically organized in the latent space. Our core motivation is to determine the rank structure of hidden states: do conflict representations concentrate along a few dominant directions, or do they disperse across many dimensions? Answering this is vital, as it dictates the optimal subspace for targeted steering interventions. To capture this geometry, we analyze the rank structure via the spectral energy of the activations.

 For each conflict type, we extract hidden states at the last non-padding token, forming matrices $\mathbf{H}_{\text{conf}}^{(l)}, \mathbf{H}_{\text{cons}}^{(l)} \in \mathbb{R}^{n \times d}$. We center each matrix as $\tilde{\mathbf{H}} = \mathbf{H} - \bar{\mathbf{H}}$ and calculate the energy ratio (ER) and its delta:
\begin{equation}
\text{ER} = \frac{\sum_{i=1}^{k} \sigma_i^2}{\|\tilde{\mathbf{H}}\|_F^2}, \;
\Delta \text{ER}_l^{(t)} = \text{ER}_{l,\text{conf}}^{(t)} - \text{ER}_{l,\text{cons}}^{(t)}
\end{equation}
Here, $\sigma_1, \ldots, \sigma_k$ represent the top-$k$ singular values (with $k{=}10$). A positive $\Delta \text{ER}$ signifies low-rank compression in dominant directions, while a negative $\Delta \text{ER}$ indicates dispersion across the broader dimensional space. Further implementation details are provided in Appendix~\ref{appendix:sea_implementation}.

\begin{figure}[t!]
  \centering
  \includegraphics[width=0.99\columnwidth]{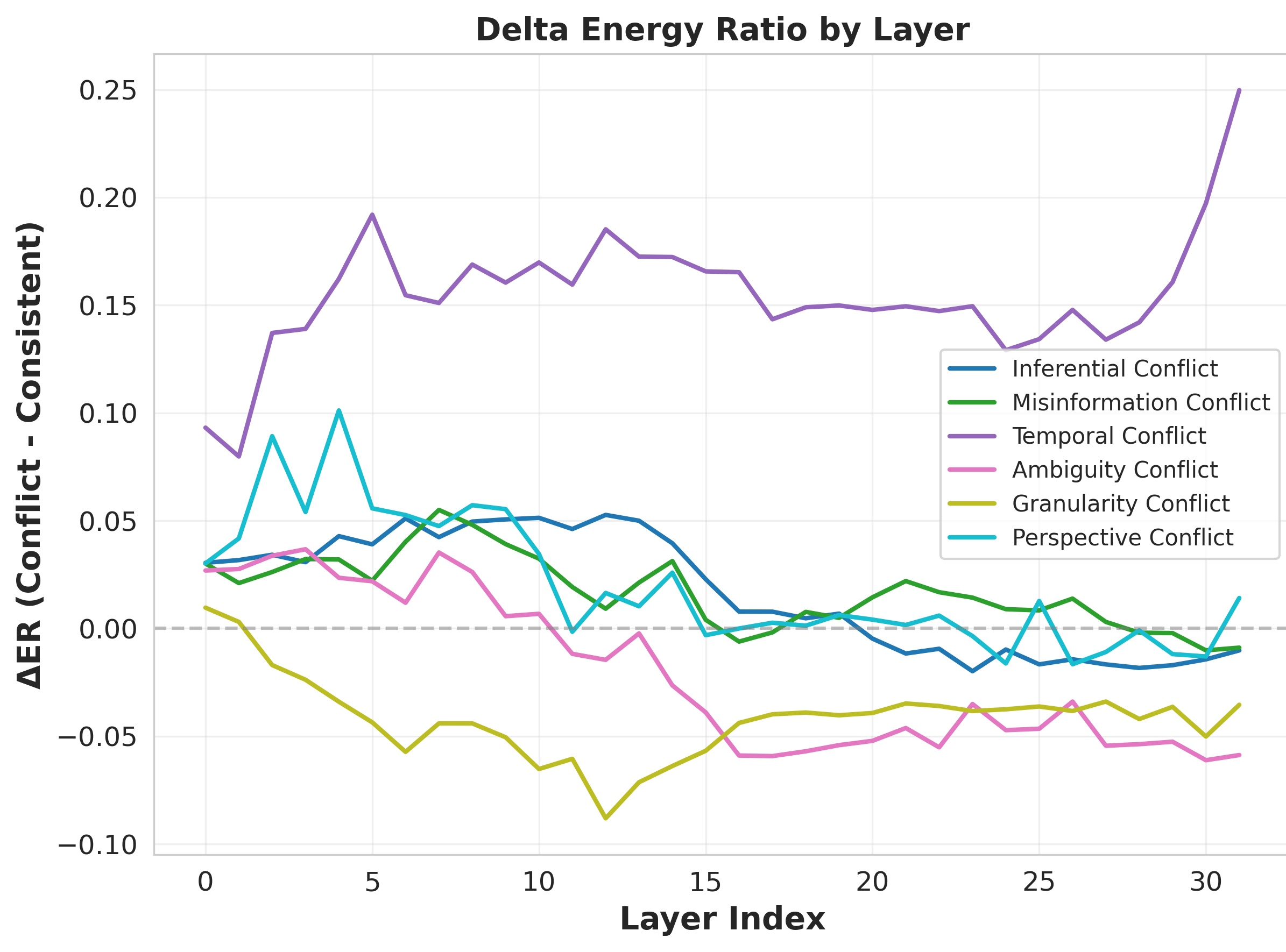}
\caption{Delta energy ratio across layers. $\Delta \text{ER} > 0$ indicates concentrated representations; $\Delta \text{ER} < 0$ indicates dispersed ones.}
  \label{fig:delta_er}
\end{figure}

The resulting geometric patterns in Figure~\ref{fig:delta_er} demonstrate that models handle different conflict types through distinct dimensional strategies. \textit{Temporal} conflicts compress into low-rank representations ($\Delta \text{ER} > 0$), effectively reducing the conflict to the singular dimension of event sequencing. Conversely, \textit{granularity} and \textit{ambiguity} conflicts disperse across dimensions ($\Delta \text{ER} < 0$); the former spans multiple levels of specificity, while the latter activates parallel lexical interpretations. \textit{Inferential} and \textit{misinformation} conflicts hover near zero, suggesting that reasoning processes reweight existing features rather than reorganizing the latent geometry. Finally, \textit{perspective} conflicts exhibit strong layer-dependence, reflecting the gradual emergence of stance as an abstract property. These distinctions indicate that the effective rank change is not merely a byproduct of conflict presence, but a diagnostic signal of how each conflict type is internally structured, represented, and resolved. Collectively, these trends are qualitatively consistent across models in Appendix~\ref{sec:sea_other_models} and reveal that conflict resolution is not a one-size-fits-all geometric process, but is instead dynamically tailored to the nature of the contradiction.

\subsection{Evidence Position Bias in Internal Representations}

We investigate "position bias" in LLMs during conflict resolution. Unlike standard retrieval tasks where bias often stems from irrelevant noise, our setting involves multiple legitimate but contradictory evidence pieces. The observed pattern therefore reveals an \emph{implicit trust allocation}: the model’s priority under direct competition.

Standard mitigations like reordering or positional adjustments are ineffective here, as they assume a single correct position or noise. Our empirical tests in Appendix~\ref{appendix:shuffling} confirm that random reordering fails to override the dominance of Evidence 1. As shown in Section~\ref{eval_result}, models disproportionately rely on the first evidence regardless of content. We analyze this systematic bias through two lenses: activation-space geometry and output-level Shapley attribution.

\begin{figure*}[ht!]
  \centering
  \includegraphics[width=0.9\textwidth]{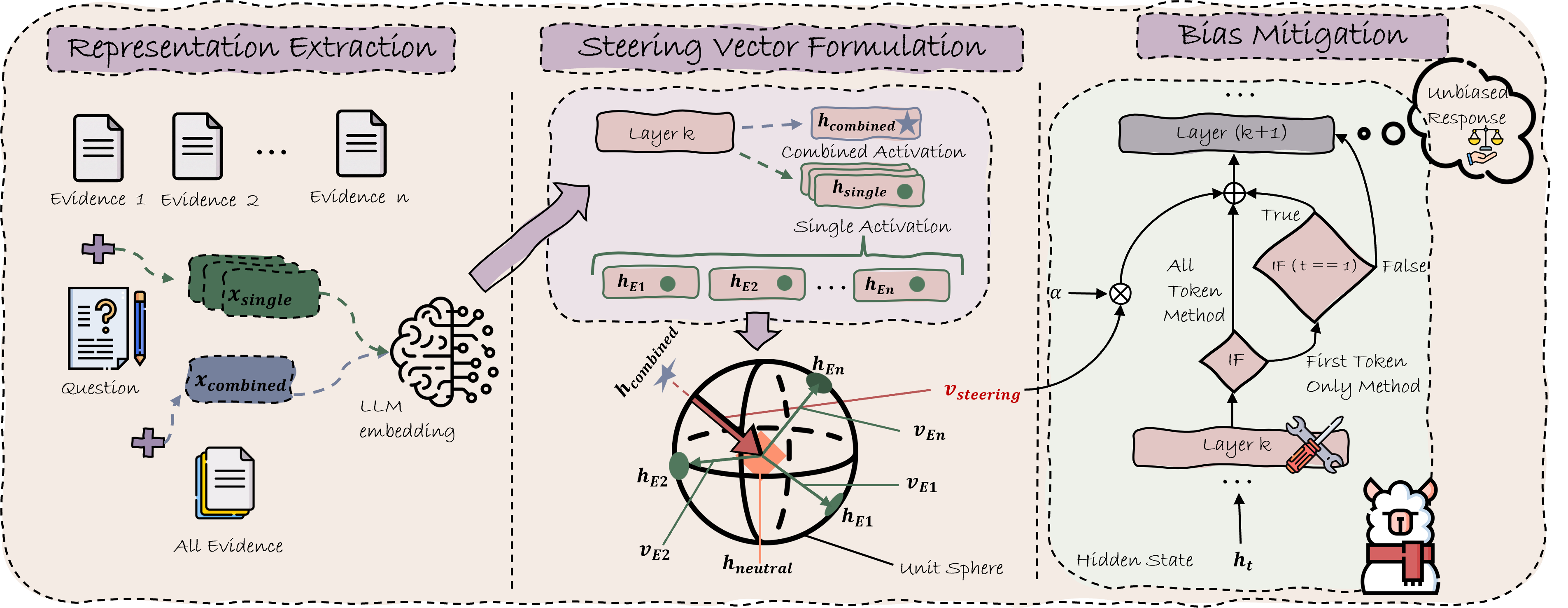}
  \caption{Three-stage activation steering method: (1) collect single-evidence and combined-evidence activations from calibration samples, (2) compute layer-wise steering directions toward neutral integration, (3) apply corrections during generation to mitigate positional bias.}
  \label{fig:method}
\end{figure*}

\noindent\textbf{Representation-Level Analysis.}
To test whether the bias originates at the representation level rather than at decoding, we compare the model's combined-evidence activation against an unbiased reference built from single-evidence activations. For a sample with $K$ pieces, we build one combined-evidence prompt and $K$ single-evidence prompts. At each layer $l$, we extract the final-token residual-stream activations: $c^{(l)}$ from the combined prompt, and $a_i^{(l)}$ from the $i$-th single-evidence prompt. If the model integrates all evidence neutrally, $c^{(l)}$ should align with the \textit{neutral center} $\mu^{(l)} = \frac{1}{K}\sum a_i^{(l)}$. We quantify this alignment using the projection:
\begin{equation}
b_i^{(l)} = \frac{(c^{(l)} - \mu^{(l)}) \cdot (a_i^{(l)} - \mu^{(l)})}{\bigl(\lVert c^{(l)} - \mu^{(l)} \rVert + \epsilon_b\bigr)\bigl(\lVert a_i^{(l)} - \mu^{(l)} \rVert + \epsilon_b\bigr)}
\end{equation}
As shown in Figure~\ref{fig:bias_simple_prompt}, this geometric bias persists across all layers. The combined representation consistently shifts toward the first evidence. This suggests the bias is a fundamental representation-level effect rather than a late-stage decoding artifact.

\noindent\textbf{Output-Level Corroboration.}
Our Shapley-based attribution framework further confirms this first-evidence dominance. As seen in Appendix Figure~\ref{fig:evidence_order_bias_pie_chart}, Evidence 1 disproportionately drives model outputs: accounting for 69.0\% of contributions in Ambiguity, 38.4\% in Granularity, and 35.0\% in Perspective. These values significantly exceed uniform baselines. Across all seven tested models, ranging from 8B to 120B parameters, Evidence 1 dominance remains a consistent, systematic artifact. This reinforces the finding that models possess an inherent architectural preference for early evidence.

\section{Activation Steering for Effective Conflict Resolution}\label{sec:method}

We propose a training-free, label-free activation steering method to mitigate position bias. Our previous analysis (Section~\ref{sec:analysis}) demonstrates that position bias is encoded as a geometric asymmetry in the residual stream. This suggests that the intervention must operate directly within the hidden-state space rather than at the input or decoding stages. If evidence were integrated uniformly, the combined-prompt activation $c^{(l)}$ would align with a position-agnostic neutral center $\mu^{(l)}$. Since the observed gap $c^{(l)} - \mu^{(l)}$ consistently points toward the first evidence, we steer the representation by translating $c^{(l)}$ back toward the neutral center. As illustrated in Figure~\ref{fig:method}, our method comprises three stages: activation collection, direction computation, and additive injection.

\noindent\textbf{Activation Collection and Direction Computation.} We partition the data into calibration (20\%) and test sets (80\%). For each calibration sample $n$, we extract the final-token residual-stream activations at layer $l$: $\{a_{n,i}^{(l)}\}$ from $K$ single-evidence prompts and $c_n^{(l)}$ from the combined prompt. We calculate the neutral center $\mu_n^{(l)} = \frac{1}{K}\sum_{i=1}^K a_{n,i}^{(l)}$ and average these values across the calibration set into $\bar{c}^{(l)}$ and $\bar{\mu}^{(l)}$. We define our steering direction as:
\begin{equation}
u^{(l)} = \frac{\bar{\mu}^{(l)} - \bar{c}^{(l)}}{\max\!\left(\|\bar{\mu}^{(l)} - \bar{c}^{(l)}\|,\,\epsilon_u\right)}
\end{equation}
This fixed direction $u^{(l)}$ is then applied to all test instances.

\noindent\textbf{Steered Inference.} At inference time, we apply the precomputed direction to nudge the residual stream toward the neutral center: $h_t^{(l)} \leftarrow h_t^{(l)} + \alpha \cdot u^{(l)}$ with $\alpha = 1.0$. We evaluate two injection schedules: \textit{First-Generated} (first step only) and \textit{All-Tokens} (every step). As shown in Table~\ref{tab:all_tasks}, both models achieve significant performance gains under All-Tokens. Llama-3.1-8B-Instruct shows improved summarization balance and substantial accuracy gains in reasoning (e.g., +19.2 points on Temporal). These improvements provide causal and targeted validation of our claim that the observed bias is rooted in the internal representation.

\noindent\textbf{Analysis and Comparisons.} To verify that the gains come from our contrastive design rather than generic context steering, we compare $u^{(l)}$ with \textit{Context-Aware Steering (CAS)}. CAS amplifies the model's overall reliance on context, whereas $u^{(l)}$ targets the evidence-integration axis identified in our analysis. Our direction wins five of six model$\times$reasoning comparisons and remains robust on summarization balance, showing that mechanistically grounded representation-level rebalancing is more effective for tasks requiring selective cross-evidence integration.

The one exception is GPT-OSS-20B Temporal, where CAS outperforms our $u^{(l)}$ (Accuracy 65.5 vs.\ 64.1). We attribute this to task-mechanism alignment: temporal reasoning requires using every timestamped piece uniformly, so the task itself rewards CAS's blanket amplification of context reliance and offers little headroom for the selective rebalancing our direction is designed for. The boundary clarifies when each direction is preferred: CAS suits tasks demanding uniform integration; our $u^{(l)}$ suits tasks requiring selective cross-evidence integration, which covers most reasoning conflicts.

The All-Tokens schedule yields the largest accuracy gains; the minor decrease in reasoning faithfulness reflects a shift toward multi-evidence reasoning that exceeds verbatim grounding.

\section{Conclusion}
We release a 5,781-sample multi-domain dataset of contextual knowledge conflicts spanning six conflict types across reasoning and summarization tasks. Our mechanistic analysis shows that LLMs often detect conflicts internally, yet still allocate disproportionate trust to earlier evidence, producing representation-level positional bias. Based on this finding, we propose a training-free, label-free activation steering method that mitigates this bias and improves evidence integration under conflict.

\section*{Limitations and Future Work}\label{sec:limitations_anchor}
We acknowledge several limitations. First, our method requires white-box access to residual-stream activations, so it applies only to open-weight models and adds inference-time overhead. Appendix~\ref{sec:evidence_order_bias_appendix} shows the same position bias in closed-source systems, so the phenomenon itself is not limited to white-box settings. Second, compute constraints restrict our steering experiments to GPT and LLaMA models at the 8B and 20B scale; scaling to 70B or 120B is left for future work. Third, our dataset uses synthetic samples for controlled manipulation; evaluating on noisy, real-world retrieval with heterogeneous evidence remains open. Fourth, our Shapley-based attribution assumes each evidence piece contributes equally in expectation, which may not hold when evidence quality varies in real-world pipelines. A weighted Shapley formulation with source-reliability priors, used alongside Faithfulness, is a promising extension.

\section*{Ethical considerations}

Our dataset contains synthetically generated misinformation designed to test model robustness against false information. The dataset does not include personally identifying information or offensive content. To mitigate potential risks of misuse, we clearly label each evidence piece as factually correct or incorrect in the dataset metadata. We emphasize that the misinformation samples are constructed solely for research purposes to evaluate conflict handling capabilities and should not be used to train models for generating misleading content. We will include explicit usage guidelines with the dataset release to prevent misuse. All base datasets used in constructing this dataset are publicly available and published through established academic channels, governed by open-access licenses that permit research reuse. 

\section*{Acknowledgments}

We thank the University of Florida Research Computing HiPerGator for providing computational resources and UF NaviGator for providing access to LLM APIs. We also acknowledge Delta at the National Center for Supercomputing Applications through allocation CIS251209 from the Advanced Cyberinfrastructure Coordination Ecosystem: Services \& Support (ACCESS) program, which is supported by U.S. National Science Foundation grants \#2138259, \#2138286, \#2138307, \#2137603, and \#2138296.

\bibliography{anthology,custom}

\appendix

\section{Extended Related Work}
\label{sec:related_work_appendix}
\paragraph{Knowledge Conflicts and Datasets.}
Large language models often produce inconsistent or hallucinatory outputs when confronted with conflicting knowledge~\cite{xie2024adaptive,xu-etal-2024-knowledge-conflicts}.
Prior datasets primarily focus on context-memory conflicts~\cite{longpre-etal-2021-entity}, often constructed through entity replacement~\cite{NEURIPS2024_baf4b960}. However, these resources exhibit four major limitations. They rely heavily on synthetic conflicts with limited real-world complexity, emphasize explicit surface-level contradictions while under-covering implicit multi-step conflicts, provide limited domain diversity, and suffer from severe class imbalance that weakens category-level analysis.
In contrast, we propose six fine-grained conflict categories spanning multiple domains, with explicit and implicit coverage across reasoning and summarization task types and type-specific implicit proportions. This design addresses key gaps in conflict complexity, diversity, and evaluation fairness. Existing datasets such as ConflictBank~\cite{NEURIPS2024_baf4b960} focus on entity-substitution-based factual conflicts, and AmbigDocs~\cite{lee2024ambigdocs} targets a single ambiguity type. To the best of our knowledge, few existing datasets jointly cover all six conflict types and two task paradigms within one multi-domain dataset.
\paragraph{Conflict Mitigation in RAG.}
Recent work addresses multi-source conflicts through several approaches. These include multi-agent deliberation \cite{li2025taming,wang2025retrievalaugmented}, conflict-driven summarization \cite{anonymous2025rethinking}, and adversarial-trained assessors \cite{choi-etal-2025-conflict,javadi2025evidencecontradictssaferretrievalaugmented}. A separate line of work uses contrastive decoding to amplify the contribution of context relative to parametric memory, most notably context-aware decoding (CAD)~\cite{shi-etal-2024-trusting}. CAD contrasts $\log p(y \mid \text{ctx})$ with $\log p(y \mid \emptyset)$, where the counterfactual baseline is the empty context. This binary contrast treats the $K$ evidence pieces as a single block: it amplifies trust in context as a whole but cannot redistribute attention or representational contribution among the individual evidence pieces. The position bias we identify is precisely an intra-context asymmetry ($b_1^{(l)} \gg b_{i>1}^{(l)}$), which CAD's reference frame cannot express; the two methods therefore address orthogonal problems and cannot be directly compared as alternatives.
These methods improve QA performance but largely treat conflict resolution as a black-box problem, often requiring additional models or training and offering limited insight into internal mechanisms.
We instead focus on internal conflict representations and propose a training-free activation steering method without external models. More importantly, to the best of our knowledge, few dataset-based studies provide a comparably broad mechanistic account of when and why LLMs fail in contextual conflict resolution.

\medskip
\paragraph{Evidence Attribution for Multi-Document Summarization.}
Shapley values have been used to quantify document importance in LLM-generated summaries \cite{ye2025fairdocumentvaluationllm}. Recent work improves efficiency through semantic clustering and decomposable utility functions \cite{ye2025fairdocumentvaluationllm,patel2025maxshapleyincentivecompatiblegenerativesearch}.
However, these methods rely on LLM-as-a-judge scoring and focus on content provider compensation scenarios rather than conflict analysis.
We introduce Shapley attribution specifically for fairness analysis in conflict scenarios, proposing the Balance metric to quantify evidence-integration bias via a normalized Gini coefficient. 

\medskip
\paragraph{Mechanistic Interpretability of LLMs.}
Prior work has probed model representations using concept activation vectors~\cite{pmlr-v80-kim18d}, spectral decomposition~\cite{NIPS2017_dc6a7e65}, and sparse autoencoders~\cite{huben2024sparse}, primarily to study factual recall, reasoning, or sentiment. Yet mechanistic analysis of how LLMs internally handle contextual knowledge conflicts remains scarce~\cite{xu-etal-2024-knowledge-conflicts}. To the best of our knowledge, few dataset-based studies jointly analyze conflict awareness, representational geometry, shared feature organization, and evidence-integration bias across conflict types and tested model architectures.

\medskip
\paragraph{Positional Bias in LLMs.}
LLMs exhibit positional bias over long contexts~\cite{liu-etal-2024-lost}, which has been addressed through positional-encoding adjustments~\cite{zhang2024found}, document reordering~\cite{jin2025from}, or training-time augmentation~\cite{NEURIPS2024_71c3451f}. However, prior studies mainly examine retrieval under noisy context, where most evidence is irrelevant and positional bias determines whether one relevant item is recovered from surrounding noise. In conflict scenarios, by contrast, all evidence is relevant but semantically competing and often logically incompatible. Positional bias therefore becomes a selective-integration problem, reflecting the model's tendency to favor specific evidence positions under genuine competition. To our knowledge, this setting remains underexplored. We provide a systematic analysis of positional bias in this regime and propose a training-free activation steering method~\cite{DBLP:journals/corr/abs-2310-01405,rimsky-etal-2024-steering} that improves reasoning performance and maintains strong summarization evidence-integration balance on tested models.

\section{Dataset Details}

\subsection{Construction Process}
\label{sec:dataset_construction}

\begin{table*}[t!]
\centering
\small
\begin{tabular}{p{3cm}p{2.0cm}p{9.6cm}}
\hline
\renewcommand{\arraystretch}{1.15} 
\textbf{Base dataset} & \textbf{Type} & \textbf{Description} \\
\hline
\renewcommand{\arraystretch}{1.00} 
AmbigDocs \\ \cite{lee2024ambigdocs} & Synthetic &
A multi-document reasoning dataset with entities sharing the same name. Tests models' ability to disambiguate entities and produce complete answers under ambiguous evidence. \\
\hline
ENTAILMENTBANK \\ \cite{dalvi-etal-2021-explaining} & Non-synthetic &
A science QA dataset with multi-step entailment chains. Shows how answers can be derived step-by-step from foundational facts. \\
\hline
NEJM-MedQA \\ \cite{Savage2024} & Non-synthetic &
Combines MedQA (USMLE-style questions) with real NEJM clinical cases. Evaluates diagnostic reasoning in both exam and real-world clinical contexts. \\
\hline
FOLIO \\ \cite{han-etal-2024-folio} & Non-synthetic &
An expert-written logical reasoning dataset annotated with first-order logic. Contains diverse reasoning structures for evaluating formal reasoning in realistic settings. \\
\hline
AllSides & Non-synthetic &
We collect recent news articles from the AllSides website. For each topic, we gather articles from media outlets representing different political stances (left, center, right), providing source material for perspective-conflict construction. \\
\hline
Perspectrum \\ \cite{chen-etal-2019-seeing} & Non-synthetic &
Contains controversial claims with stance sentences and supporting evidence. Designed for multi-perspective discovery and understanding. \\
\hline
ROAST-ABSA \\ \cite{chebolu2024roastreviewlevelopinionaspect} & Non-synthetic &
A review-level aspect-based sentiment analysis dataset across multiple languages and domains. Includes aspect-opinion-sentiment-target quadruples for fine-grained analysis. \\
\hline
SciFact \\ \cite{wadden-etal-2020-fact} & Non-synthetic &
A scientific claim verification dataset with expert-labeled evidence as supporting or refuting. Evaluates evidence retrieval, verification, and explanation. \\
\hline
ConflictBank \\ \cite{NEURIPS2024_baf4b960} & Synthetic &
A systematically constructed dataset that injects plausible but incorrect distracting knowledge. Tests model robustness and truthfulness under conflicting evidence. \\
\hline
CONFLICTS \\ \cite{cattan2025draggedconflictsdetectingaddressing} & Non-synthetic &
A RAG dataset built on real web search results with expert-annotated conflict types. Evaluates conflict recognition and response quality under multi-source knowledge conflicts. \\
\hline
\end{tabular}
\caption{Base datasets used to construct our dataset.}
\label{tab:base_datasets}
\end{table*}

\begin{figure*}[t]
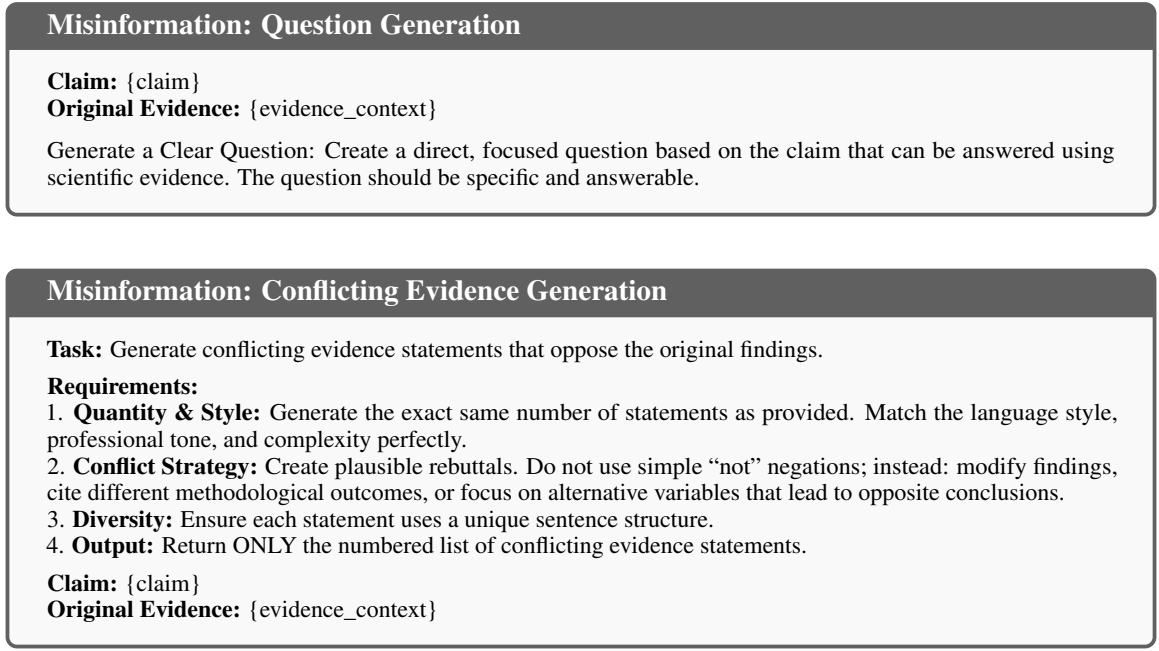

  \centering
  \begin{tcolorbox}[
    colback=gray!5!white,
    colframe=gray!75!black,
    fonttitle=\bfseries,
    fontupper=\small,
    title=Misinformation: Question Generation,
    rounded corners,
    width=0.95\textwidth
  ]
  \textbf{Claim:} \{claim\}\\
  \textbf{Original Evidence:} \{evidence\_context\}\\[6pt]
  Generate a Clear Question: Create a direct, focused question based on the claim that can be answered using scientific evidence. The question should be specific and answerable.
  \end{tcolorbox}
  \vspace{6pt}

  \begin{tcolorbox}[
    colback=gray!5!white,
    colframe=gray!75!black,
    fonttitle=\bfseries,
    fontupper=\small,
    title=Misinformation: Conflicting Evidence Generation,
    rounded corners,
    width=0.95\textwidth
  ]
  \textbf{Task:} Generate conflicting evidence statements that oppose the original findings.\\[4pt]
  \textbf{Requirements:}\\
  1. \textbf{Quantity \& Style:} Generate the exact same number of statements as provided. Match the language style, professional tone, and complexity perfectly.\\
  2. \textbf{Conflict Strategy:} Create plausible rebuttals. Do not use simple ``not'' negations; instead: modify findings, cite different methodological outcomes, or focus on alternative variables that lead to opposite conclusions.\\
  3. \textbf{Diversity:} Ensure each statement uses a unique sentence structure.\\
  4. \textbf{Output:} Return ONLY the numbered list of conflicting evidence statements.\\[4pt]
  \textbf{Claim:} \{claim\}\\
  \textbf{Original Evidence:} \{evidence\_context\}
  \end{tcolorbox}
  \caption{Prompts for misinformation conflict construction.}
  \label{fig:prompt_misinformation}
\end{figure*}

\begin{figure*}[t]
  \centering
  \includegraphics[width=0.9\textwidth]{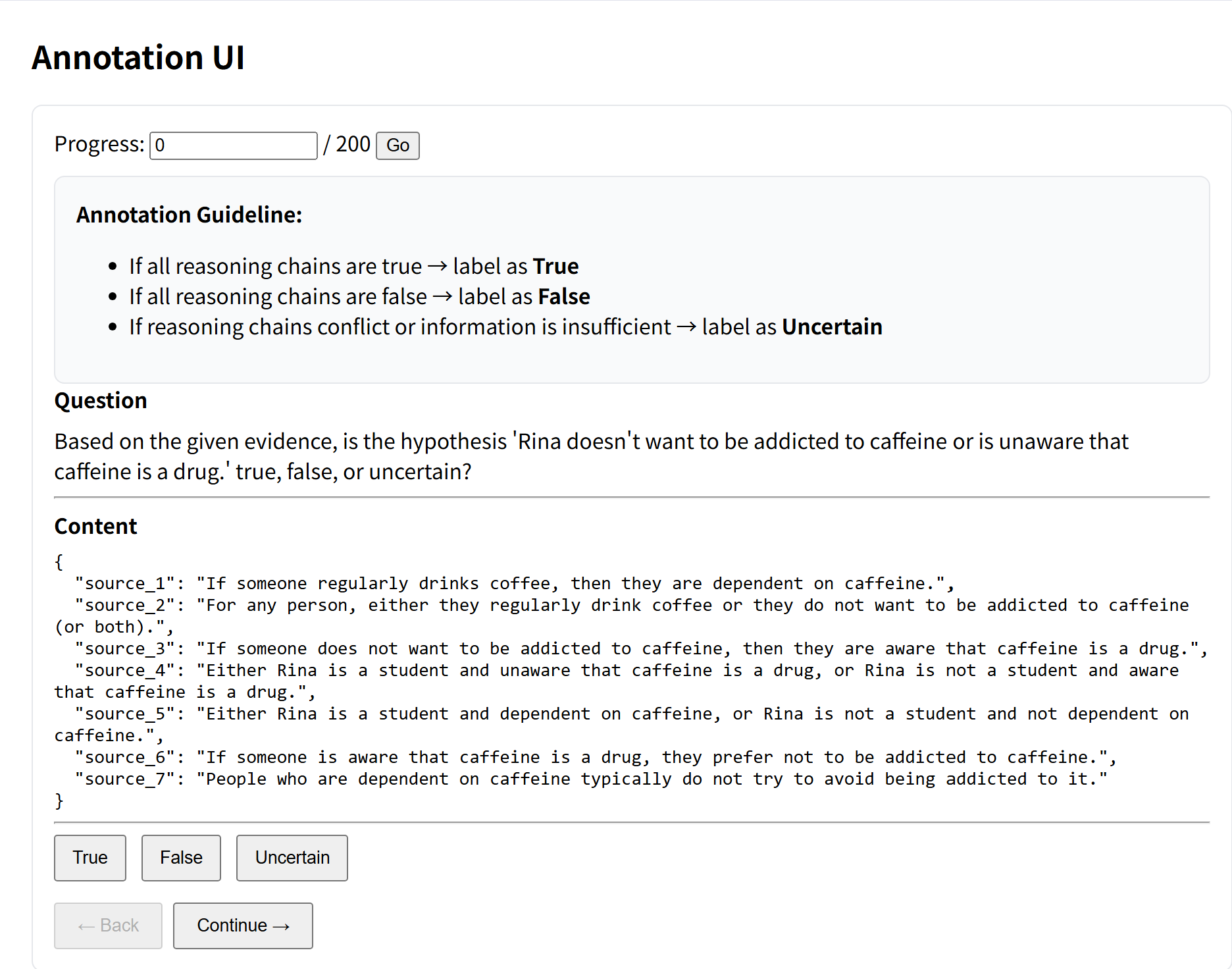}
  \caption{Annotation interface for FOLIO manual verification. Annotators select True/False if reasoning chains lead to a unique answer, or Uncertain if chains produce conflicting conclusions.}
  \label{fig:folio_ann_ui}
\end{figure*}

Table~\ref{tab:base_datasets} lists the ten base datasets used to construct our dataset.
The construction procedure for each conflict type is described in Section~\ref{sec:construction}; here we provide the GPT-5 prompts used for semi-synthetic data generation.
Figure~\ref{fig:prompt_misinformation} shows the two-step prompt for misinformation conflicts (SciFact subset): question generation followed by conflicting evidence generation.
Figure~\ref{fig:prompt_folio} shows the prompt for inferential conflicts (FOLIO and ENTAILMENTBANK subsets), which augments original premises with conflicting reasoning branches.
The annotation interface for manual re-annotation of FOLIO instances is shown in Figure~\ref{fig:folio_ann_ui}.
Figure~\ref{fig:prompt_temporal} shows the prompt for temporal conflicts (ConflictBank subset), which attaches explicit temporal markers to generate time-dependent questions.
Figure~\ref{fig:prompt_conflictbank_general} shows the prompt for generating factual questions for other ConflictBank conflict types.


\begin{figure*}[t]
  \centering
  \begin{tcolorbox}[
    colback=gray!5!white,
    colframe=gray!75!black,
    fonttitle=\bfseries,
    fontupper=\small,
    title=FOLIO: Conflicting Premise Generation,
    rounded corners,
    width=0.95\textwidth
  ]
  Your task:\\
  1. \textbf{Decision:} With 50\% probability, choose either to:\\
  \quad - \textbf{Option A:} Add 1--2 plausible statements that create a conflicting reasoning branch (requires at least one inference step; avoid trivial ``X vs.\ not-X'').\\
  \quad - \textbf{Option B:} Keep the premises as they are.\\
  2. \textbf{Output:} Return ONLY the final numbered list of statements.\\
  3. \textbf{Constraints:} No meta-talk, no ``source\_\#'' references, no explanations.\\[4pt]
  \textbf{Premises:} \{premises\_raw\_json\}\\
  \textbf{Hypothesis:} \{conclusion\}
  \end{tcolorbox}
  \caption{Prompt for FOLIO inferential conflict construction.}
  \label{fig:prompt_folio}
\end{figure*}

\begin{figure*}[t]
  \centering
  \begin{tcolorbox}[
    colback=gray!5!white,
    colframe=gray!75!black,
    fonttitle=\bfseries,
    fontupper=\small,
    title=Temporal Conflicts: Question Generation,
    rounded corners,
    width=0.95\textwidth
  ]
  Based on two conflicting claims about the same subject, generate ONE concise open-ended question with a temporal marker.\\[4pt]
  \textbf{Inputs:} Subject, default claim, temporal claim, default evidence, temporal evidence\\[4pt]
  \textbf{Requirements:}\\
  1. Extract ALL timestamps/dates from both evidences.\\
  2. Identify the attribute being described (e.g., employer, award, title).\\
  3. Randomly choose to ask about a SPECIFIC YEAR before, between, or after identified timestamps.\\
  4. Use patterns like: ``What was \{subject\}'s [attribute] in [year]?'' or ``Who was \{subject\}'s [attribute] as of [year]?''\\
  5. Prefer specific years over time ranges.\\
  6. Do NOT use yes/no questions or multiple-choice formats.\\
  7. Self-contained, one sentence, $\leq$ 20 words.\\
  8. Output only the question (no explanations, no metadata).
  \end{tcolorbox}
  \caption{Prompt for temporal conflict question generation.}
  \label{fig:prompt_temporal}
\end{figure*}

\begin{figure*}[t]
  \centering
  \begin{tcolorbox}[
    colback=gray!5!white,
    colframe=gray!75!black,
    fonttitle=\bfseries,
    fontupper=\small,
    title=General ConflictBank: Question Generation,
    rounded corners,
    width=0.95\textwidth
  ]
  Given a subject with a default claim (assumed correct) and a conflicting claim, generate ONE concise factual question about the subject without revealing either claim explicitly.\\[4pt]
  \textbf{Inputs:} Subject, default claim, conflicting claim\\[4pt]
  \textbf{Requirements:}\\
  - Use simple WH-questions like ``Where did \{subject\} work?'', ``When was \{subject\} born?'', ``Who is \{subject\} married to?''\\
  - Do NOT include claim contents in the question.\\
  - Self-contained, one sentence.
  \end{tcolorbox}
  \caption{Prompt for general ConflictBank question generation.}
  \label{fig:prompt_conflictbank_general}
\end{figure*}

\subsection{Decision Rules for Borderline Conflict-Type Overlaps}
\label{sec:decision_rules}

Our taxonomy is defined by operational, structural criteria rather than surface intuition. We document the decision rule for each pair of conflict types whose surface descriptions could otherwise be confused.

\paragraph{Temporal vs. Misinformation.} The criterion is whether each evidence piece holds true at some time point. In \textit{temporal} conflicts, every evidence piece is true at its own time point; the conflict arises because claim validity changes over time. In \textit{misinformation} conflicts, some evidence is factually incorrect, and each piece carries an accuracy label in the dataset metadata.

\paragraph{Ambiguity vs. Inferential.} The criterion is the task goal, not merely the number of reasoning steps. In \textit{ambiguity} conflicts, each evidence piece describes a different real entity that shares the same name (e.g., ``Jordan'' can refer to the basketball player or to a university teacher with the same name); each piece is understandable on its own, no cross-evidence reasoning is needed, and the goal is to integrate and present every entity fairly rather than let the more popular referent dominate. Ambiguity instances are identified directly from the entity-disambiguation metadata of the source dataset. In \textit{inferential} conflicts, the conflict is not limited to entities: it only emerges after combining multiple pieces of evidence through reasoning, and the goal is to derive the single correct conclusion.

\paragraph{Perspective vs. Granularity.} The criterion is compatibility. \textit{Perspective} conflicts involve incompatible stances on the same question. \textit{Granularity} conflicts involve compatible answers at different levels of specificity.

\subsection{Data Samples}
\label{app:data_samples}

We provide representative examples for each conflict type in our dataset.
Figure~\ref{fig:inferential_conflict} illustrates an inferential conflict where logical reasoning leads to an uncertain conclusion.
Figure~\ref{fig:misinformation_conflict} presents a misinformation conflict involving contradictory scientific claims with accuracy labels.
Figure~\ref{fig:temporal_conflict} demonstrates a temporal conflict requiring temporal reasoning across events.
Figure~\ref{fig:ambiguity_conflict} shows an ambiguity conflict where one name refers to multiple distinct entities.
Figure~\ref{fig:granularity_conflict} provides a granularity conflict example with varying levels of detail,
and Figure~\ref{fig:perspective_conflict} highlights a perspective conflict involving diverse viewpoints on a policy issue.

\begin{figure*}[ht!]
\begin{tcolorbox}[colback=blue!5!white, colframe=blue!75!black, title=Inferential Conflict Example, width=\textwidth]
\textbf{Question:} Based on the given evidence, is the hypothesis 'Miroslav Venhoda loved music.' true, false, or uncertain? The final answer should be 'True', 'False', or 'Uncertain'. Only answer 'Uncertain' if the reasoning results from different evidence pieces conflict with each other.

\textbf{Evidence 1:} Miroslav Venhoda was a Czech choral conductor.

\textbf{Evidence 2:} Miroslav Venhoda specialized in the performance of Renaissance and Baroque music.

\textbf{Evidence 3:} All choral conductors are musicians.

\textbf{Evidence 4:} Some musicians love music.

\textbf{Evidence 5:} In 1946, Miroslav Venhoda published a book titled Method of Studying Gregorian Chant.

\textbf{Evidence 6:} Authors of instructional books on musical practice often have strong enthusiasm for music.

\textbf{Evidence 7:} A scholar's focus on liturgical chant can be motivated by historical or religious interest rather than a personal love of music.

\textbf{Ground Truth:} Uncertain
\end{tcolorbox}
\caption{Example of an Inferential Conflict, demonstrating logical uncertainty.}
\label{fig:inferential_conflict}
\end{figure*}

\begin{figure*}[ht!]
\begin{tcolorbox}[colback=red!5!white, colframe=red!75!black, title=Misinformation Conflict Example, width=\textwidth]
\textbf{Question:} In a bleomycin-induced mouse model of pulmonary fibrosis, does pharmacological activation of AMPK (e.g., with metformin) reduce established lung fibrosis in an AMPK-dependent manner? Answer with Yes or No.

\textbf{Evidence 1:}
In a bleomycin model of lung fibrosis in mice, metformin therapeutically accelerates the resolution of well-established fibrosis in an AMPK-dependent manner.

\textbf{Evidence 2:}
In primary myofibroblasts isolated from IPF lungs, pharmacological activation of AMPK did not attenuate fibrotic activity; instead, it was accompanied by diminished mitochondrial biogenesis and persistence of apoptosis resistance.

\textbf{Evidence 3:}
Contrary to expectations, these investigations indicate that heightened AMPK signaling underlies non-resolving, pathologic fibrotic remodeling, and argue that metformin (and related AMPK agonists) are unlikely to reverse established fibrosis because they maintain myofibroblasts in an activated, apoptosis-resistant state.

\textbf{Accuracy Labels:} [True, False, False]

\textbf{Ground Truth:} Yes
\end{tcolorbox}
\caption{Example of a Misinformation Conflict with specific accuracy labels for each evidence.}
\label{fig:misinformation_conflict}
\end{figure*}

\begin{figure*}[ht!]
\begin{tcolorbox}[colback=green!5!white, colframe=green!75!black, title=Temporal Conflict Example, width=\textwidth]
\textbf{Question:} Which university did Alvaro Mata attend in 2030?

\textit{Note: Evidence snippets are truncated for readability; we use [...] as the truncation marker.}

\textbf{Evidence 1:}
In the heart of Cleveland, Ohio[...] lies Cleveland State University. It is here that Alvaro Mata, a researcher of unyielding curiosity and dedication, embarked on a transformative journey[...] It is the early 2000s, and Cleveland State University is thriving[...] Alvaro Mata emerges as a bright and ambitious individual[...] Evidence of Alvaro's attendance at Cleveland State University can be found in the university's library archives, where academic papers and theses reveal his prolific output as a student researcher.

\textbf{Evidence 2:}
Renowned Researcher Alvaro Mata Celebrates 14-Year Milestone at Macquarie University. Sydney, Australia - December 15, 2038. Today, Macquarie University in Sydney marked a significant milestone as esteemed researcher Alvaro Mata completed an impressive 14-year tenure at the institution[...] "It's hard to believe it's been 14 years since I first stepped onto this campus," Mata reflected.

\textbf{Ground Truth:} Macquarie University
\end{tcolorbox}
\caption{Example of a Temporal Conflict requiring chronological reasoning.}
\label{fig:temporal_conflict}
\end{figure*}

\begin{figure*}[ht!]
\begin{tcolorbox}[colback=orange!5!white, colframe=orange!75!black, title=Ambiguity Conflict Example, width=\textwidth]
\textbf{Question:} What is Peter Cusack known for?

\textbf{Evidence 1:}
Peter Cusack (musician) is an artist and musician who is a member of CRiSAP (Creative Research in Sound Arts Practice), and is a research staff member at the London College of Communication. He was a founding member and director of the London Musicians' Collective. He is best known as a member of the avant garde musical quartet, Alterations (1978–1986; with Steve Beresford, David Toop, and Terry Day), and the creator of field and wildlife recording-based albums.

\textbf{Evidence 2:}
Peter Cusack made his premiership début with the Sydney Roosters in the 1998 NRL season. A front-rower, Cusack was one of the last remaining top level league players to hold a job outside football, working part-time as a plumber. He played in the 2000 NRL Grand Final loss to the Brisbane Broncos. Cusack was awarded the 2002 Sydney Roosters season's "Clubman of the Year" and played in their 2002 NRL Grand Final victory over the New Zealand Warriors.

\end{tcolorbox}
\caption{Example of an Ambiguity Conflict involving two different individuals with the same name.}
\label{fig:ambiguity_conflict}
\end{figure*}

\begin{figure*}[ht!]
\begin{tcolorbox}[colback=purple!5!white, colframe=purple!75!black, title=Granularity Conflict Example, width=\textwidth]
\textbf{Task:} Summarize the following evidence in 2-3 sentences.

\textbf{Evidence 1:}
I'm only here for taste and texture, and the recent cups haven't hit the same mark. The broth turns thin unless I dump the entire seasoning in, and even then I reach for soy sauce to wake it up. The noodles rehydrate fine—springy enough after five minutes under a kettle—but the base lacks that savory backbone it used to have[...] If you're chasing a rich sip straight from the cup, this version feels watered down.

\textbf{Evidence 2:}
The "lower sodium" splash on the front sent me straight to the Nutrition Facts, and the numbers tell a different story than I expected. The serving size is smaller than my older cups[...] Net weight is down, the water line looks lower, and the sodium per serving drops mostly because the serving itself is lighter. It reads like less soup rather than a smarter recipe.

\textbf{Evidence 3:}
As an overall lunch option, these are still easy: boil water, fill to the line, lid on for five minutes, stir hard, and you've got something warm at your desk. The flavor is dependable once you learn your routine[...] The convenience is great for busy days, yet the front-of-pack claims don't line up neatly with what I taste and how much I'm actually eating.

\textbf{Evidence 4:}
I prefer the large McDougall cups because one can carry me through an afternoon of meetings without snacks. Filled to the line, it's closer to a bowl than a snack cup, and sometimes I can't finish it in one sitting.

\textbf{Evidence 5:}
For anyone sensitive to salt, using about three-quarters of the seasoning packet has been the sweet spot. I stir in the dry mix gradually, taste, and stop right before it tips into that heavy-salty zone; a squeeze of lemon and some scallions round it out.

\end{tcolorbox}
\caption{Example of a Granularity Conflict where evidence varies in detail and focus.}
\label{fig:granularity_conflict}
\end{figure*}

\begin{figure*}[ht!]
\begin{tcolorbox}[colback=yellow!5!white, colframe=yellow!75!black, title=Perspective Conflict Example, width=\textwidth]
\textbf{Question:} Should the school day be extended?

\textbf{Evidence 1:}
Evidence from Chicago Public Schools shows that assigning underprepared 9th graders to a double-period (extended time) algebra course raised math achievement, increased credit accumulation, and led to higher graduation and college enrollment rates—indicating targeted extensions of the school day for core instruction can produce substantial gains for struggling students.

\textbf{Evidence 2:}
The AAP recommends middle and high schools start at 8:30 a.m. or later because adolescents need 8–10 hours of sleep; insufficient sleep is linked to worse grades, depression, and accident risk. Extending the school day without shifting to later start times may exacerbate sleep deprivation and harm learning and health.

\textbf{Evidence 3:}
National survey data show teachers report substantially higher job-related stress and burnout than other working adults, with time pressure and workload as leading contributors. Lengthening the school day risks worsening burnout and turnover unless paired with added staffing, planning time, and compensation.

\textbf{Evidence 4:}
On school days, juvenile crime and victimization rates peak in the hours immediately after school (roughly 2–6 p.m.). Extending the school day to provide structured supervision during these hours is argued to reduce delinquency and enhance community safety.

\textbf{Evidence 5:}
A package of reforms that included substantially increased instructional time produced large gains in math and reading. The study suggests extended time can be effective when embedded in a broader model, implying time alone is unlikely to deliver similar results.

\textbf{Evidence 6:}
The EEF estimates extending school time yields a small average impact on learning (around +2 months' progress per year) at moderate cost; benefits are larger when time is tightly targeted (e.g., tutoring), raising questions about the cost-effectiveness of blanket school-day extensions.

\end{tcolorbox}
\caption{Example of a Perspective Conflict featuring diverse viewpoints on school hour extensions.}
\label{fig:perspective_conflict}
\end{figure*}

\section{Evaluation Details}

\subsection{Evaluation via Different Scorer Models}
\label{sec:eval_diff_mode}
Our Balance metric relies on a scorer model to compute the value function. As defined in Eq.~2 of Section~\ref{sec:evaluation_metrics}, this function maps each evidence subset to a normalized utility score. To verify the robustness of our metric, we test two different scorer models: Llama-3.1-8B and Gemma-2B. These models differ in both scale and pretraining approach.

Table~\ref{tab:bal_robustness_two_scorers} shows Balance scores computed using both scorer models. While absolute values differ between scorers, relative performance patterns remain highly consistent across tested models and conflict types. Crucially, model rankings by Balance remain nearly identical across scorers. For instance, \texttt{gemini-2.5-pro} consistently achieves the best Balance scores in Ambiguity under both scorers, while \texttt{llama-3.1-8b-instruct} consistently shows the highest Balance scores across conflict types. These results suggest that our Shapley-based Balance metric is robust to scorer choice and less likely to be driven by scorer-specific artifacts.

\begin{table*}[t]
\centering
\small
\setlength{\tabcolsep}{5pt}
\renewcommand{\arraystretch}{1.1}
\newcolumntype{Y}{>{\centering\arraybackslash}X}
\begin{tabularx}{\textwidth}{l *{6}{Y}}
\toprule
\textbf{Model} &
\multicolumn{2}{c}{\textbf{Ambiguity}} &
\multicolumn{2}{c}{\textbf{Granularity}} &
\multicolumn{2}{c}{\textbf{Perspective}} \\
\cmidrule(lr){2-3}\cmidrule(lr){4-5}\cmidrule(lr){6-7}
& \textbf{Llama-3.1-8B} & \textbf{Gemma-2B}
& \textbf{Llama-3.1-8B} & \textbf{Gemma-2B}
& \textbf{Llama-3.1-8B} & \textbf{Gemma-2B} \\
\midrule
gpt-5                     & 24.7 & 20.7 & 16.7 & 15.2 & 26.3 & 19.9 \\
claude-4.5-sonnet         & 23.6 & 19.4 & 18.9 & 15.7 & 26.2 & 18.1 \\
gemini-2.5-pro            & 19.3 & 15.9 & 20.0 & 16.3 & 27.5 & 17.3 \\
gpt-oss-120b              & 30.0 & 25.1 & 20.0 & 15.6 & 27.3 & 22.0 \\
gpt-oss-20b               & 38.4 & 27.0 & 18.5 & 16.5 & 26.7 & 17.0 \\
llama-3.1-70b-instruct    & 26.3 & 20.5 & 24.6 & 23.0 & 30.2 & 26.1 \\
llama-3.1-8b-instruct     & 39.9 & 27.7 & 29.9 & 28.5 & 33.1 & 26.8 \\
\bottomrule
\end{tabularx}
\caption{Balance scores under two different scorer models. The relative rankings and performance patterns remain consistent across scorers, demonstrating robustness of the Shapley-based Balance metric to scorer choice.}
\label{tab:bal_robustness_two_scorers}
\end{table*}

\subsection{Human Annotation and Metric Validation}
\label{human_eval}
\begin{figure*}[ht!]
  \centering
  \includegraphics[width=\textwidth]{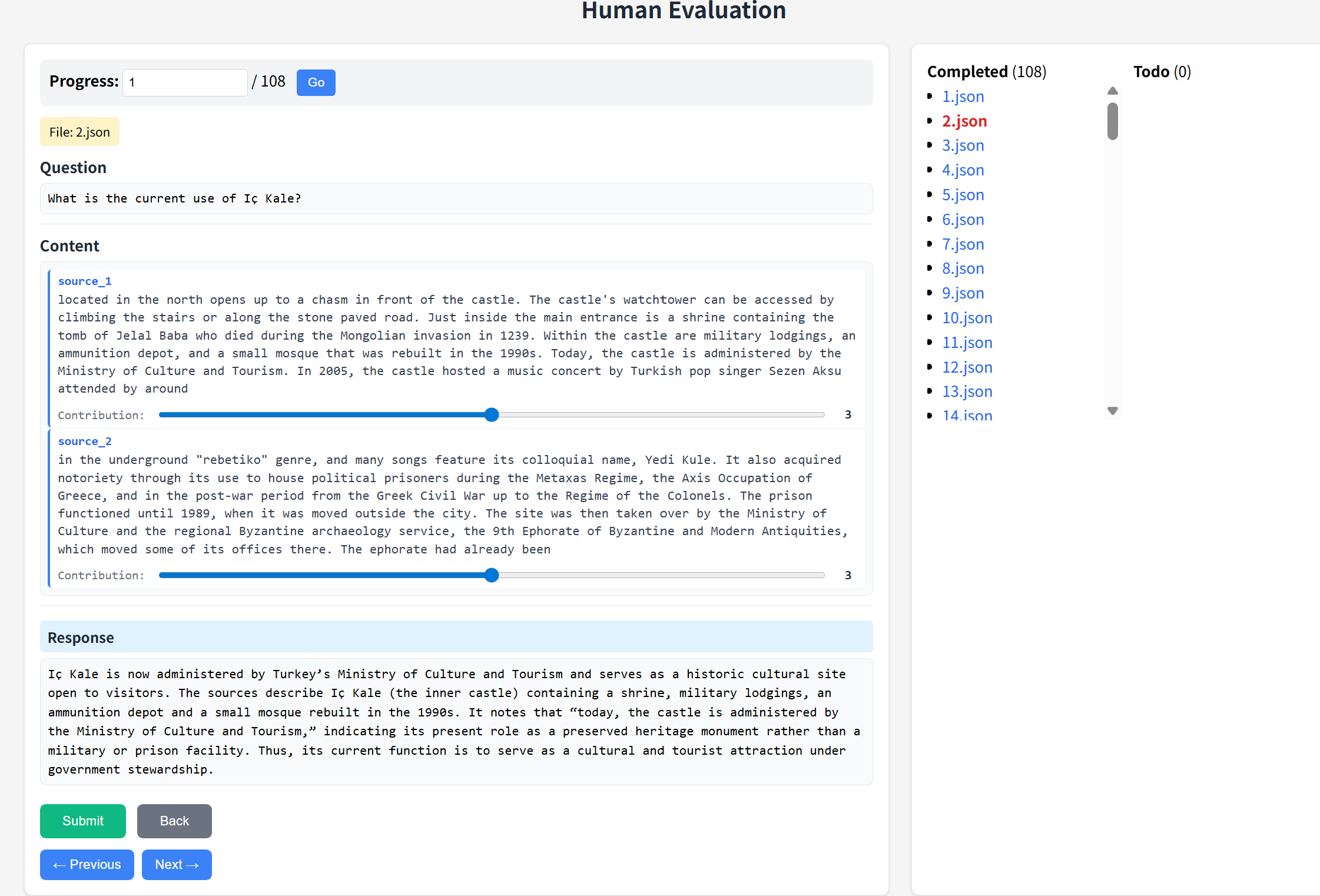}
  \caption{Annotation interface for rating evidence contributions on a 1--5 scale.}
  \label{fig:human_eval_ui}
\end{figure*}

To validate our Shapley-based metric, we conduct a human annotation study.
Two independent annotators evaluate evidence contributions for a subset of samples from summarization tasks.
Figure~\ref{fig:human_eval_ui} shows the annotation interface.
Annotators rate each evidence source's contribution to the response on a 1--5 scale.
We provide detailed guidelines in Figure~\ref{fig:annotation_guidelines}.
Table~\ref{tab:kappa_agreement} reports Cohen's Kappa coefficients across 108 annotated samples.
Human annotators achieve moderate agreement ($\kappa = 0.5199$).
Our automatic metric also agrees with both annotators ($\kappa = 0.5208$ and $0.5014$).
These results suggest that the metric is broadly consistent with human judgments of evidence contribution. Table~\ref{tab:human_evaluation} reports per-model Balance scores from both human raters and the LLM-as-a-Judge across the three summarization tasks.

\begin{table}[ht]
\centering
\scriptsize
\setlength{\tabcolsep}{3pt}
\renewcommand{\arraystretch}{1.0}
\begin{tabular}{lccc}
\toprule
\textbf{Model} & \textbf{Ambig.} & \textbf{Gran.} & \textbf{Persp.} \\
\midrule
\multicolumn{4}{l}{\textit{Human Evaluation}} \\
claude-4.5-sonnet & \textbf{21.18} & \textbf{16.53} & \textbf{12.85} \\
gpt-oss-120b      & \underline{22.51} & \underline{17.32} & \underline{13.97} \\
llama-3.1-8b      & 28.80 & 23.89 & 15.91 \\
\addlinespace[3pt]
\midrule
\addlinespace[3pt]
\multicolumn{4}{l}{\textit{LLM-as-a-Judge}} \\
claude-4.5-sonnet & \textbf{20.62} & \textbf{9.06} & \textbf{17.89} \\
gpt-oss-120b      & \underline{24.66} & \underline{13.15} & \underline{18.30} \\
llama-3.1-8b      & 30.99 & 21.02 & 21.44 \\
\bottomrule
\end{tabular}
\caption{Balance scores (normalized Gini \%, lower is better) from human evaluation and LLM-as-a-Judge on three models spanning diverse capability levels.}
\label{tab:human_evaluation}
\end{table}

\begin{figure*}[t]
  \centering
  \begin{tcolorbox}[
    enhanced,
    colback=gray!5!white,
    colframe=gray!75!black,
    fonttitle=\bfseries,
    fontupper=\small,
    title=Evidence Contribution Annotation Guidelines,
    rounded corners,
    width=0.95\linewidth
  ]
  
  \textbf{Task Overview:} Evaluate the contribution of each evidence source to the model's response for a given question. Rate each source on a 1--5 scale.\\[6pt]
  
  \textbf{Rating Scale:}\\
  \textbf{1 -- Minimal Contribution:} Evidence barely reflected in response; removing it would not affect core content.\\
  \textbf{2 -- Minor Contribution:} Some information mentioned but peripheral; plays supporting role.\\
  \textbf{3 -- Moderate Contribution:} Key information clearly reflected; important for response completeness.\\
  \textbf{4 -- Major Contribution:} Core content prominently featured; significant impact on main arguments.\\
  \textbf{5 -- Dominant Contribution:} Response primarily derived from this evidence; plays decisive role.\\[6pt]
  
  \textbf{Core Principles:}\\
  \textbf{1. Focus on relative distribution, not absolute values.} Ratings should reflect \emph{relative contribution differences} between sources.\\
  \textit{Example A (Significant disparity):} Source 1: 5, Source 2: 1 $\rightarrow$ Response heavily relies on Source 1.\\
  \textit{Example B (Balanced):} Source 1: 3, Source 2: 3 $\rightarrow$ Both sources contribute equally.\\
  \textit{Example C (Low but balanced):} Source 1: 1, Source 2: 1 $\rightarrow$ Response relies on model's own knowledge.\\[4pt]
  
  \textbf{2. Base judgment on content matching:}\\
  -- \textit{Direct quotation:} Response uses verbatim or paraphrased content from evidence.\\
  -- \textit{Information correspondence:} Facts, data, viewpoints in response traceable to evidence.\\
  -- \textit{Structural influence:} Response organization influenced by evidence structure.\\[4pt]
  
  \textbf{3. Evaluate each evidence independently,} then use score distribution to reflect relative relationships.\\[6pt]
  
  \textbf{Annotation Steps:}\\
  1. Read the \textbf{Question} to understand the core inquiry.\\
  2. Read all \textbf{Sources} to understand the information each provides.\\
  3. Read the \textbf{Response} carefully, analyzing its content and structure.\\
  4. \textbf{Rate each source:} Identify which parts of the response come from each evidence, assess the weight of that information, and assign a 1--5 score.\\
  5. \textbf{Check consistency:} Ensure score distribution reasonably reflects relative contributions.\\[6pt]
  
  \textbf{Important Notes:}\\
  -- \textit{Avoid position bias:} Do not favor evidence appearing earlier.\\
  -- \textit{Distinguish quality from contribution:} High-quality evidence unused in response should receive a low score.\\
  -- \textit{Handle conflicting evidence:} Focus on which evidence's viewpoint dominates in the response.
  
  \end{tcolorbox}
  \caption{Annotation guidelines for evaluating evidence contribution to model responses.}
  \label{fig:annotation_guidelines}
\end{figure*}

\section{Mechanistic Analysis Details}

\subsection{Conflict-Consistent Pair Construction and Cross-Validation Statistics}
\label{sec:consistent_pairs}

The concept-vector analysis in Section~\ref{sec:analysis} relies on pairing each conflict instance $x_{\text{conf}}^{(t)}$ with a consistent counterpart $x_{\text{cons}}^{(t)}$ in which the evidence no longer produces a conflict. We construct each consistent counterpart by preserving the question and surrounding context and replacing or filtering only the evidence set, following per-type rules:

\begin{itemize}
\item \textbf{Misinformation.} We retain only the evidence pieces labeled factually correct in the source annotation and discard the conflicting ones, so that every remaining piece supports the same gold conclusion.
\item \textbf{Inferential (FOLIO and EntailmentBank).} We keep the original entailment chain and remove the GPT-5-generated conflicting branch, so that all premises jointly support the original hypothesis label.
\item \textbf{Temporal.} We anchor on a single timestamp and retain only the evidence pieces consistent with that anchor; we introduce no new content.
\item \textbf{Granularity.} We keep evidence pieces drawn from the same diagnostic specificity level (either all broad-syndrome or all specific-disease), without mixing levels.
\item \textbf{Perspective.} We retain evidence pieces that share the same stance label and discard those expressing the opposing stance.
\item \textbf{Ambiguity.} We retain evidence pieces that refer to a single entity, using the source dataset's disambiguation metadata.
\end{itemize}

Each conflict instance is paired with exactly one consistent counterpart, and we use these conf-versus-cons pairs as the labeled inputs to the linear probes in Section~\ref{sec:analysis}. Table~\ref{tab:cv_stats} reports per-type sample counts and 5-fold stratified cross-validation statistics. Probes are linear logistic-regression classifiers with inverse-regularization strength $C{=}1.0$ and a maximum of 1000 iterations, fit on residual-stream activations at each layer.

\begin{table}[ht]
\centering
\footnotesize
\setlength{\tabcolsep}{4pt}
\begin{tabular}{lccc}
\toprule
\textbf{Conflict type} & $|x_{\text{conf}}|$ & $|x_{\text{cons}}|$ & \textbf{Fold (train / val)} \\
\midrule
Misinformation & 1004 & 1004 & 1606 / 402 \\
Inferential    &  787 &  787 & 1259 / 315 \\
Temporal       &  960 &  960 & 1536 / 384 \\
Granularity    & 1020 & 1020 & 1632 / 408 \\
Perspective    & 1010 & 1010 & 1616 / 404 \\
Ambiguity      & 1000 & 1000 & 1600 / 400 \\
\bottomrule
\end{tabular}
\caption{Conflict-consistent pair counts and 5-fold stratified cross-validation statistics per conflict type. Each conflict instance is paired with one consistent counterpart, so $|x_{\text{conf}}| = |x_{\text{cons}}|$ by construction; per-fold training and validation sizes are computed from the combined pool $|x_{\text{conf}}| + |x_{\text{cons}}|$ with an 80/20 stratified split.}
\label{tab:cv_stats}
\end{table}

\subsection{Concept Vector Analysis on Additional Models and Implicit Conflicts}
\label{sec:cv_additional_models_implicit}

We extend concept vector analysis to additional models and examine implicit versus explicit conflicts.

\begin{figure}[ht!]
  \centering
  \includegraphics[width=0.95\columnwidth]{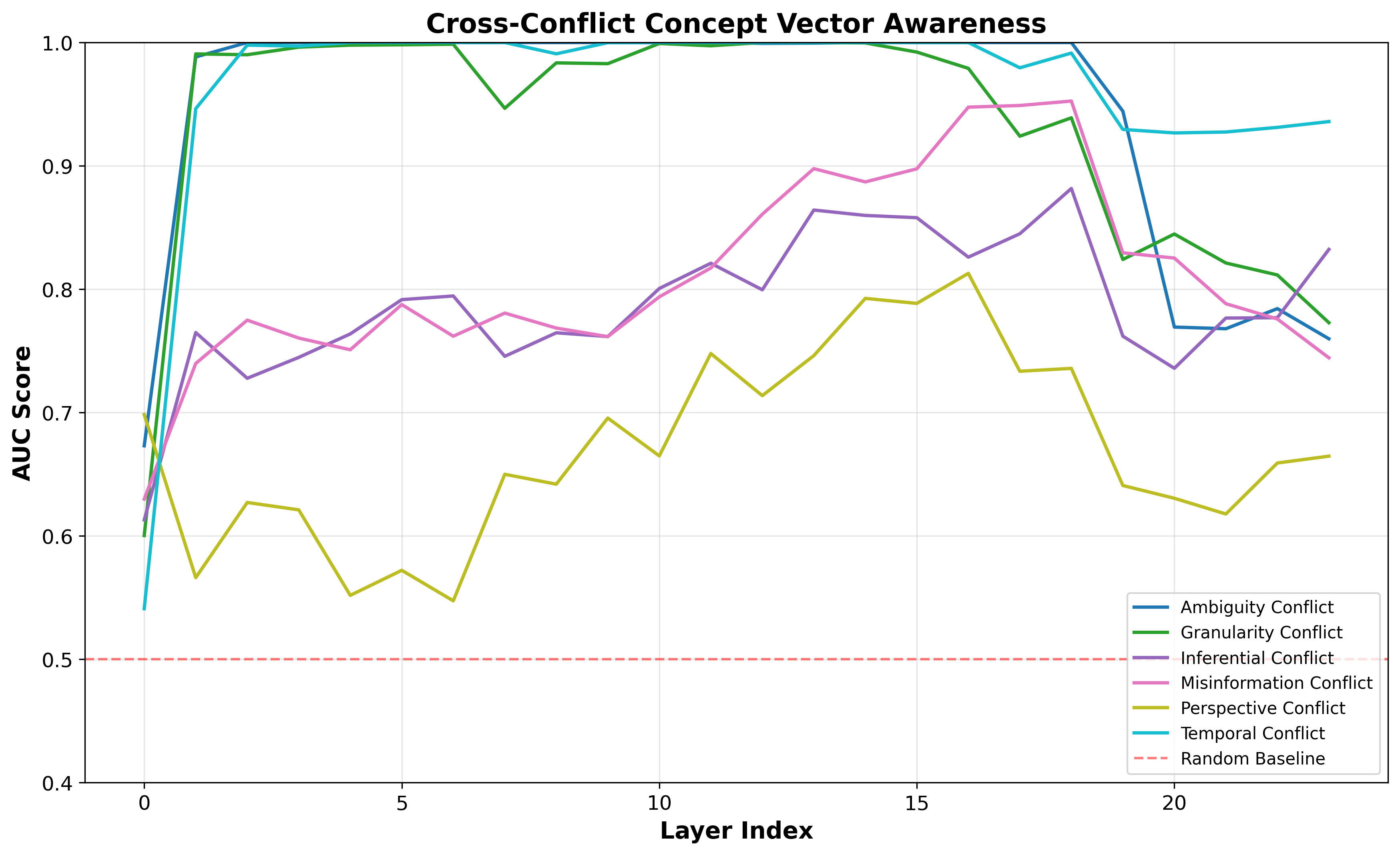}
  \caption{Layer-wise AUC for conflict awareness in GPT-OSS-20B. Despite fewer layers, the model demonstrates consistent awareness patterns across conflict types.}
  \label{fig:auc_comparison_gpt20b}
\end{figure}

\paragraph{Analysis on Additional Models.}
 Figure~\ref{fig:auc_comparison_gpt20b} shows that GPT-OSS-20B exhibits awareness trends similar to those of Llama models despite having fewer layers. Temporal and ambiguity conflicts reach saturation quickly, while other conflict types show gradual emergence. The model exhibits strong conflict awareness across all types. These findings suggest that our observations generalize across the tested model scales and architectures.

\paragraph{Implicit vs. Explicit Conflicts.}
We analyze perspective and inferential conflicts from different source datasets. We operationalize the explicit/implicit distinction using the structural criterion in Section~\ref{sec:dataset_overview}. A conflict is \textbf{explicit} if it requires at most one inferential step to detect, and \textbf{implicit} if it requires at least two cross-evidence inferential steps.

\textbf{Inferential conflicts.}
ENTAILMENTBANK (explicit): Each reasoning chain presents a step-by-step entailment that directly yields a stated conclusion. When a second evidence piece asserts an incompatible conclusion, the contradiction is identifiable by direct comparison of their final claims, requiring a single inferential step.
FOLIO (implicit): GPT-5-generated premises introduce a conflicting reasoning branch by altering conditional or logical dependencies. To identify the conflict, a reader must trace causal relationships through multiple premises and compare derivations across evidence chains. This process requires combining at least two evidence pieces through conditional logic before the incompatibility surfaces, satisfying our implicit criterion.

\textbf{Perspective conflicts.}
Perspectrum (explicit): Evidence pieces contain explicit stance sentences that directly affirm or negate the same claim (e.g., ``X is beneficial'' vs.\ ``X is harmful''). The viewpoint conflict is identifiable by direct comparison of these surface propositions in a single inferential step.
AllSides (implicit): Evidence pieces describe the same event through selective emphasis, differential fact selection, and divergent rhetorical framing, without any single statement directly contradicting another. Detecting the underlying viewpoint conflict requires integrating implicit stances across multiple documents. The process demands multi-step cross-document synthesis to identify what each article implies but does not state, which satisfies our implicit criterion on structural grounds independent of the data source.

\begin{figure}[ht!]
  \centering
  \includegraphics[width=0.95\columnwidth]{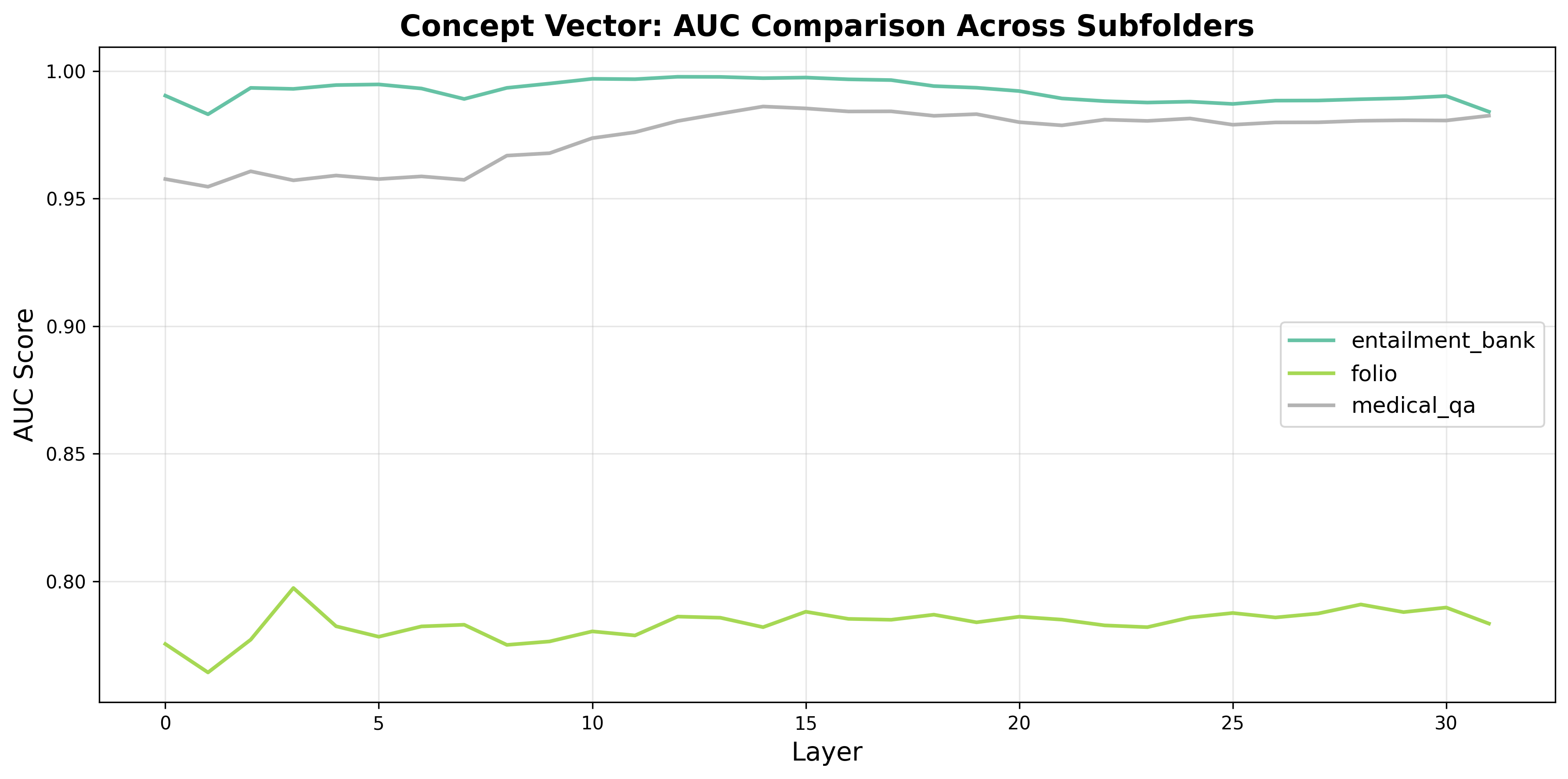}
  \caption{Concept vector AUC comparison for inferential conflicts across datasets. EntailmentBank (explicit) shows consistently higher AUC than FOLIO (implicit), indicating stronger awareness of surface-level logical conflicts.}
  \label{fig:subfolder_auc_comparison_infer}
\end{figure}

Figure~\ref{fig:subfolder_auc_comparison_infer} compares inferential conflicts. ENTAILMENTBANK (explicit) maintains consistently high AUC throughout all layers. FOLIO (implicit) shows lower AUC in early layers and continues to decline in final layers. The gap suggests that models rely heavily on surface-level signals for conflict detection and struggle when contradiction requires multi-step cross-evidence inference.

\begin{figure}[ht!]
  \centering
  \includegraphics[width=0.95\columnwidth]{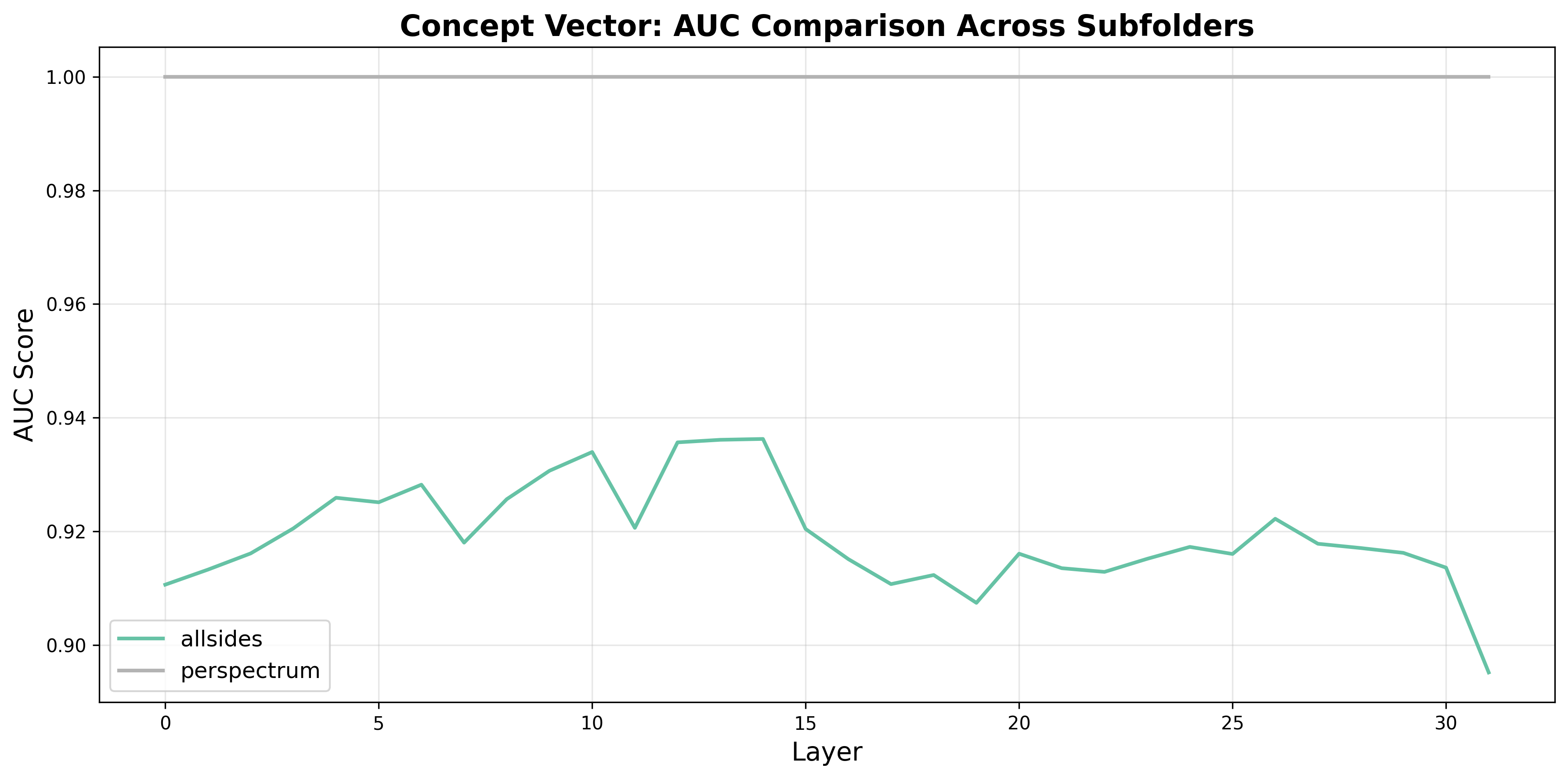}
  \caption{Concept vector AUC comparison for perspective conflicts across datasets. Perspectrum (explicit) achieves stable high AUC, while AllSides (implicit) fluctuates near chance level, indicating reduced sensitivity to subtle viewpoint differences.}
  \label{fig:subfolder_auc_comparison_per}
\end{figure}

Figure~\ref{fig:subfolder_auc_comparison_per} compares perspective conflicts. The difference is even more striking: Perspectrum (explicit) achieves stable high AUC across layers, whereas AllSides (implicit) fluctuates near chance level. This suggests that model representations are less sensitive to conflicts that require multi-step cross-document synthesis, regardless of whether those conflicts arise from logical structure (FOLIO) or selective framing (AllSides). The shared structural factor is the number of inferential steps required.

\subsection{Spectral Energy Analysis Implementation Notes}
\label{appendix:sea_implementation}

The three-stage pipeline and the ER and $\Delta$ER formulas are defined in Section~\ref{sec:spectral_energy}. We list here the numerical and software details omitted from the main text. We compute the top-$k$ singular values with PyTorch's \texttt{svd\_lowrank} on the centered activation matrix $\tilde{\mathbf{H}}$. We set $k{=}10$ and $\epsilon_{\mathrm{ER}}=10^{-12}$ for numerical stability. Activations are taken at the last non-padding token of each sample, matching the protocol used in Section~\ref{sec:spectral_energy}.

\subsection{Spectral Energy Analysis in Other Models}
\label{sec:sea_other_models}

\begin{figure}[ht!]
  \centering
  \includegraphics[width=\columnwidth]{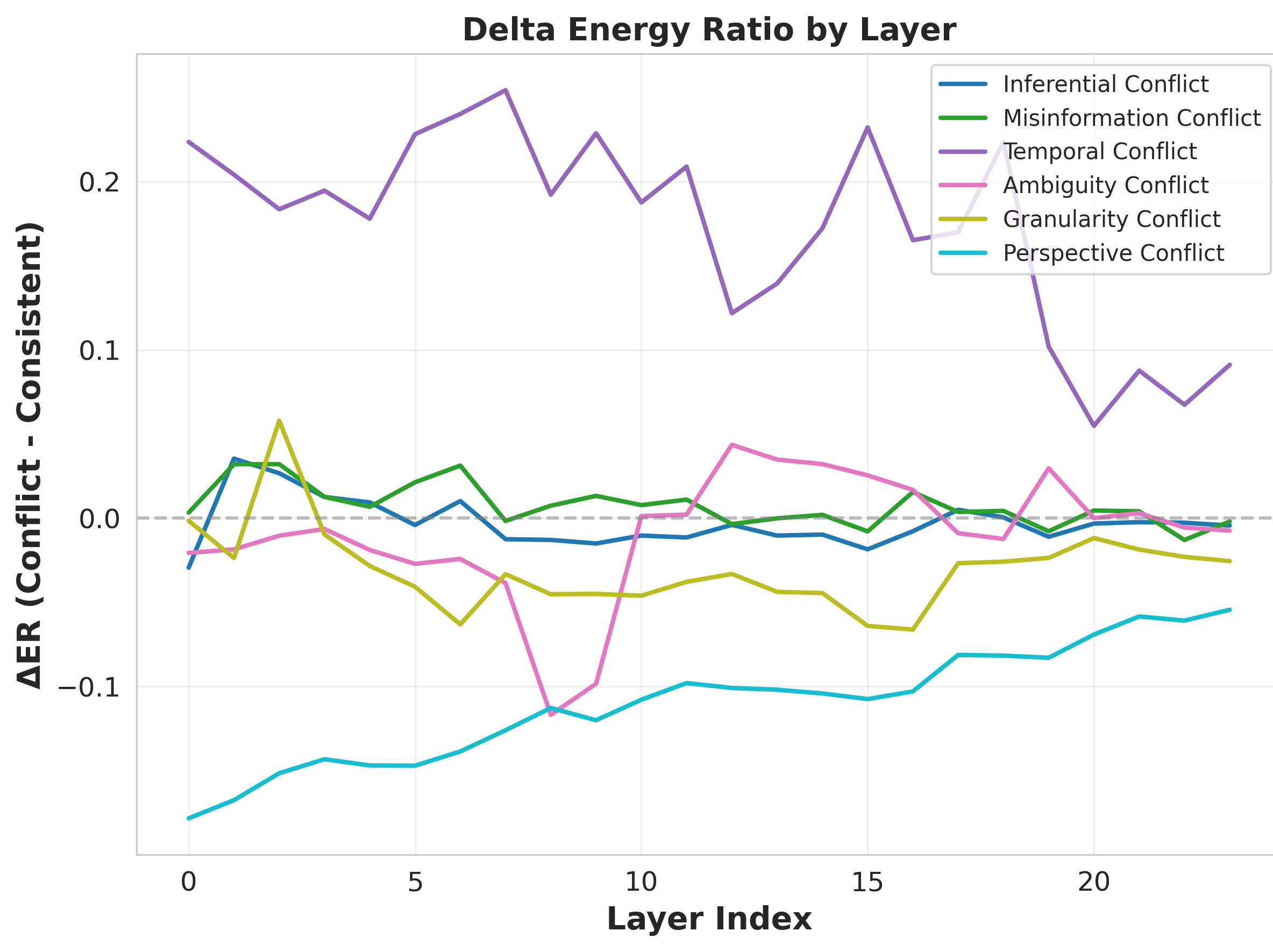}
  \caption{Delta energy ratio in \texttt{openai/gpt-oss-20b} across layers. $\Delta \text{ER} > 0$ indicates conflict states are more concentrated (low-rank); $\Delta \text{ER} < 0$ indicates more dispersed (high-rank).}
  \label{fig:delta_er_gpt20b}
\end{figure}

Figure~\ref{fig:delta_er_gpt20b} shows delta energy-ratio patterns in \texttt{openai/gpt-oss-20b}. The type-specific geometric patterns remain consistent with Llama-3.1-8B, suggesting that similar geometric trends appear across tested models.

\subsection{Evidence Position Bias Across Tested Models}
\label{sec:evidence_order_bias_appendix}

\begin{figure*}[ht]
  \centering
  \includegraphics[width=\textwidth]{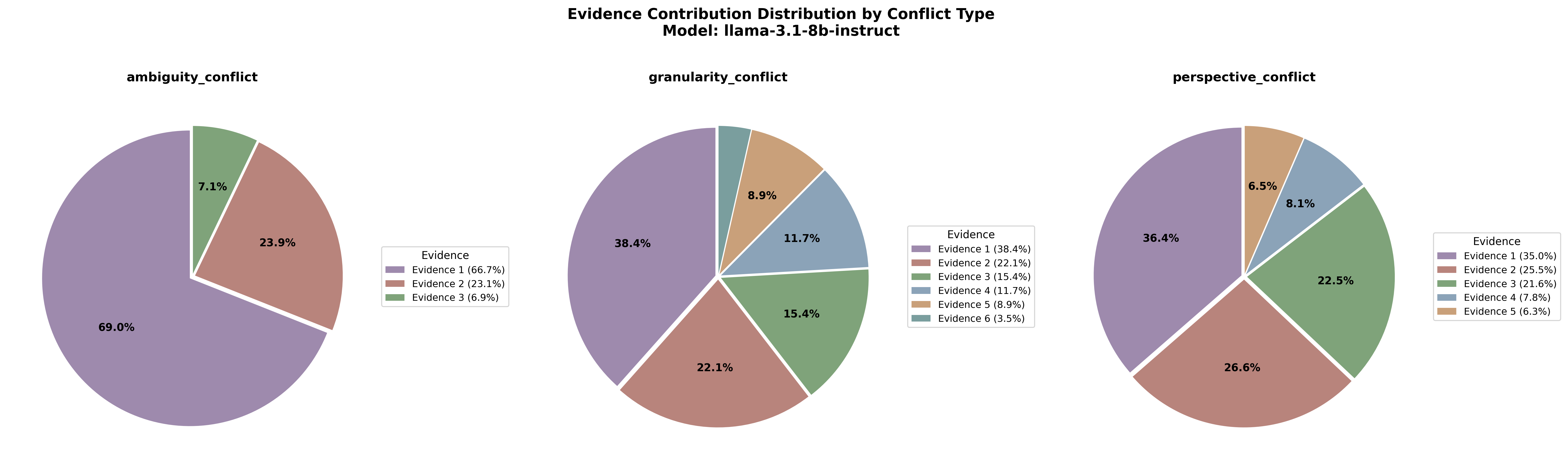}
  \caption{Evidence contribution distribution from output-level analysis. As described in Section~\ref{sec:evaluation_metrics}, the distribution is highly non-uniform, with earlier evidence often dominating.}
  \label{fig:evidence_order_bias_pie_chart}
\end{figure*}

\begin{figure*}[ht]
  \centering
  \includegraphics[width=\textwidth]{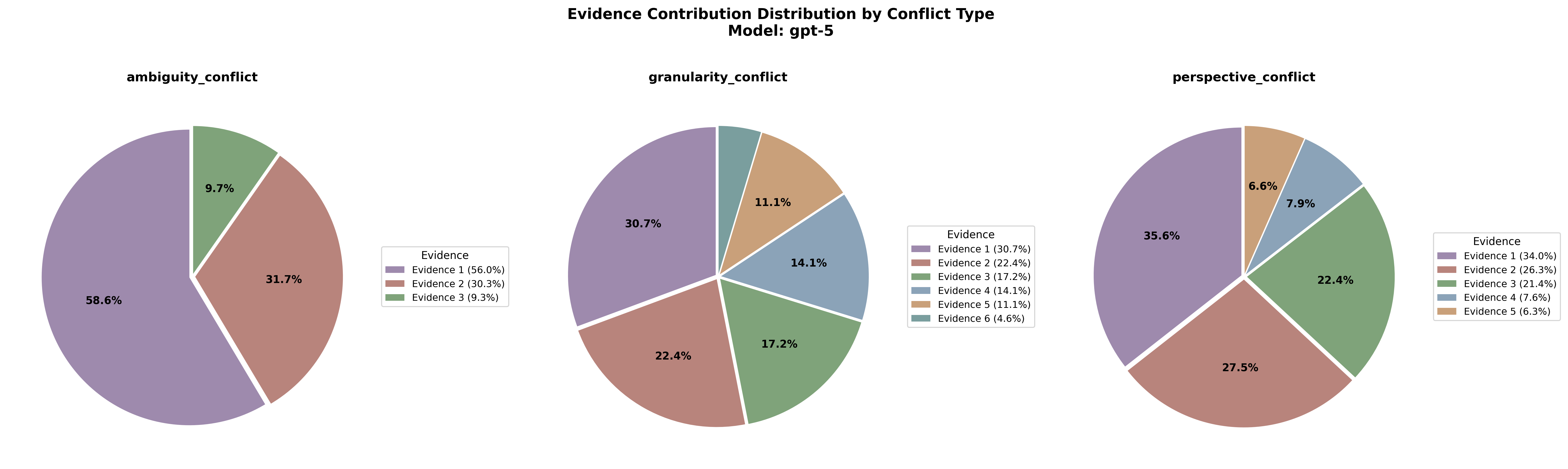}
  \caption{Evidence position bias (pie chart) for \texttt{gpt-5}.}
  \label{fig:evidence_order_bias_pie_chart_gpt5}
\end{figure*}

\begin{figure*}[ht]
  \centering
  \includegraphics[width=\textwidth]{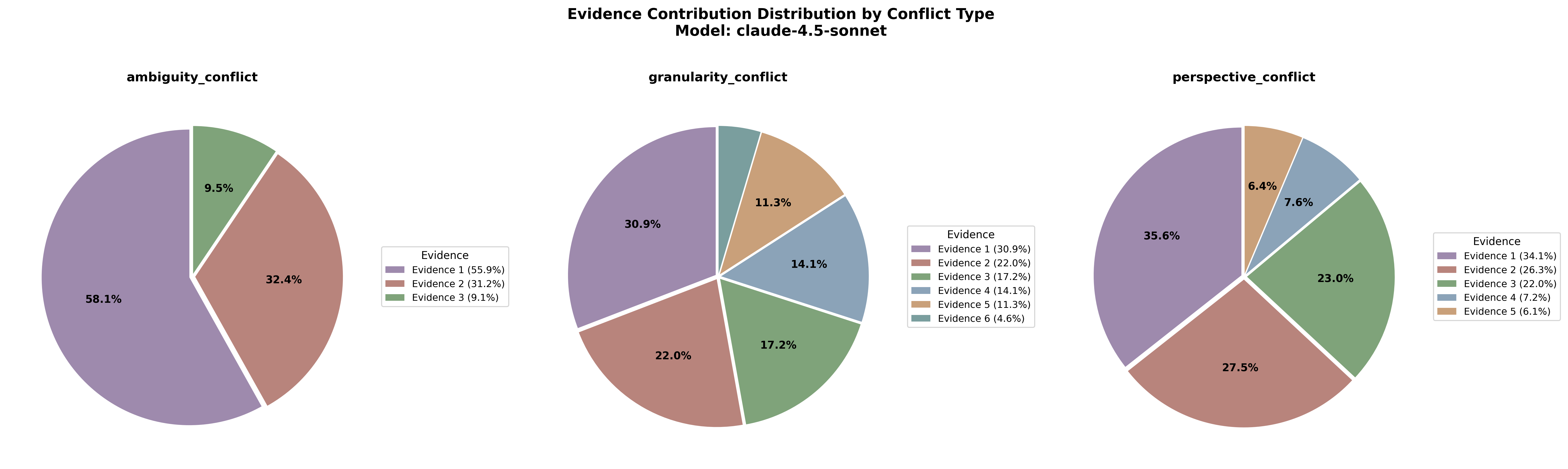}
  \caption{Evidence position bias (pie chart) for \texttt{claude-4.5-sonnet}.}
  \label{fig:evidence_order_bias_pie_chart_claude45}
\end{figure*}

\begin{figure*}[ht]
  \centering
  \includegraphics[width=\textwidth]{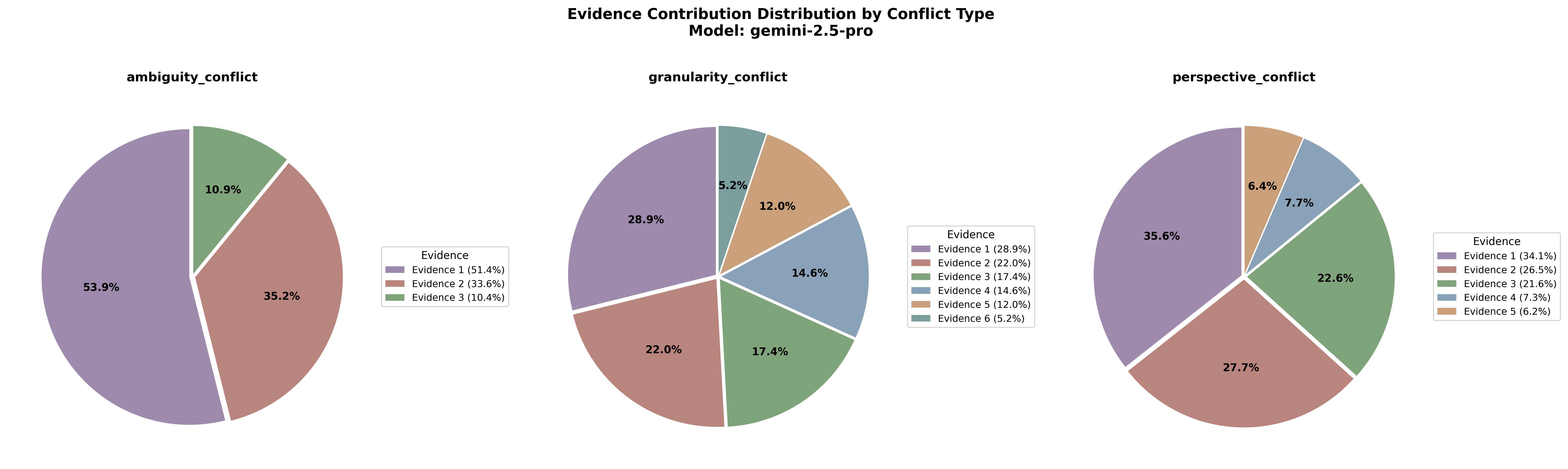}
  \caption{Evidence position bias (pie chart) for \texttt{gemini-2.5-pro}.}
  \label{fig:evidence_order_bias_pie_chart_gemini25}
\end{figure*}

\begin{figure*}[ht]
  \centering
  \includegraphics[width=\textwidth]{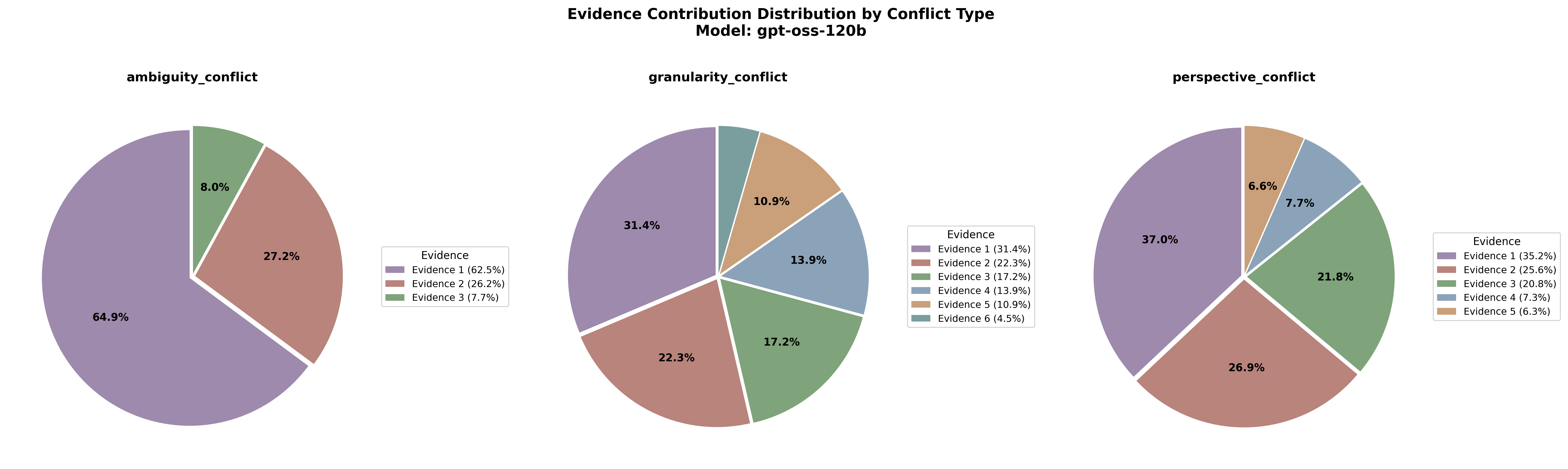}
  \caption{Evidence position bias (pie chart) for \texttt{gpt-oss-120b}.}
  \label{fig:evidence_order_bias_pie_chart_gptoss120b}
\end{figure*}

\begin{figure*}[ht]
  \centering
  \includegraphics[width=\textwidth]{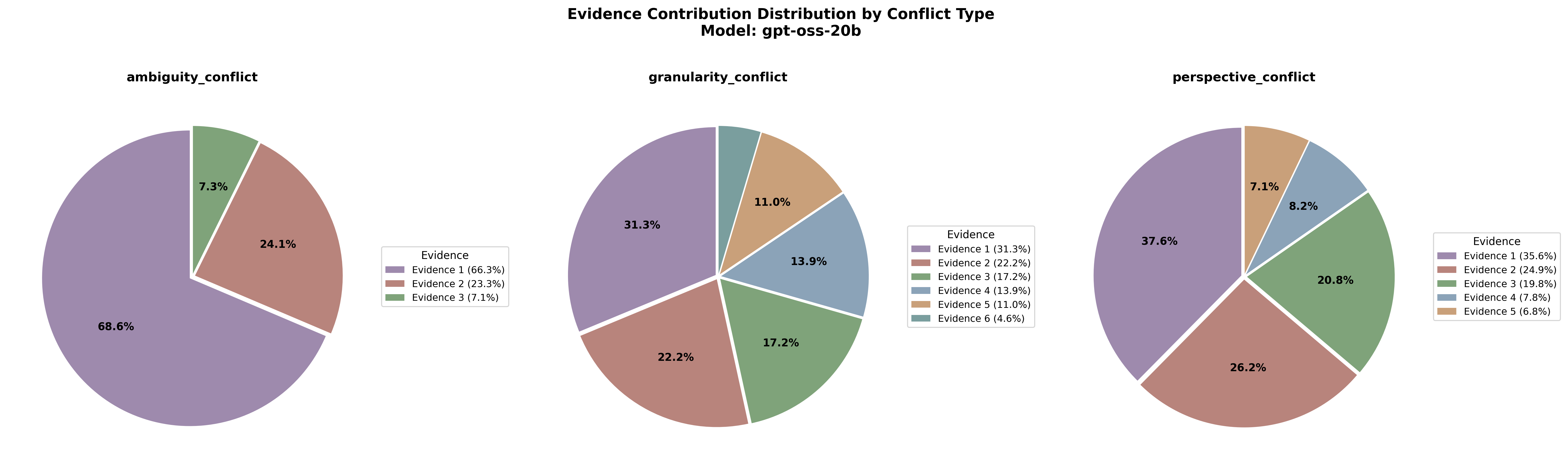}
  \caption{Evidence position bias (pie chart) for \texttt{gpt-oss-20b}.}
  \label{fig:evidence_order_bias_pie_chart_gptoss20b}
\end{figure*}

\begin{figure*}[ht]
  \centering
  \includegraphics[width=\textwidth]{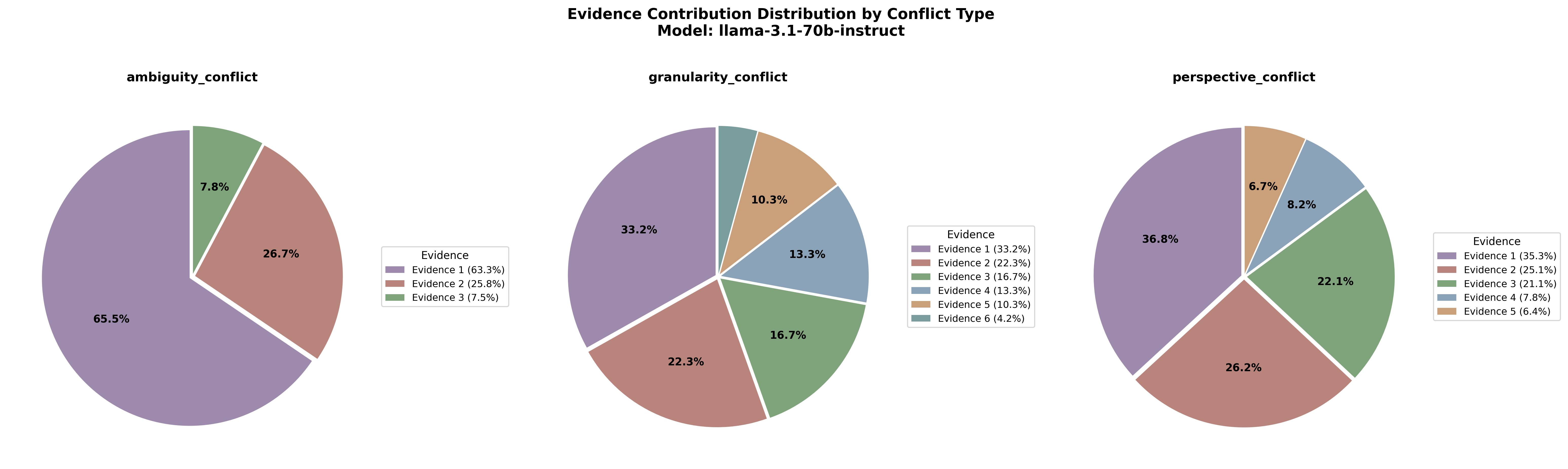}
  \caption{Evidence position bias (pie chart) for \texttt{llama-3.1-70b-instruct}.}
  \label{fig:evidence_order_bias_pie_chart_llama3170b}
\end{figure*}

We analyze evidence-position bias across seven tested models spanning different scales and training approaches.
Figures~\ref{fig:evidence_order_bias_pie_chart}--\ref{fig:evidence_order_bias_pie_chart_llama3170b} show evidence-contribution distributions for these tested models. The bias is pervasive within this dataset: earlier evidence consistently receives larger contributions. This pattern appears in both small models (Llama-3.1-8B) and large models (GPT-5, Claude-4.5-Sonnet), and in both proprietary and open-source systems. In our experiments, this bias is not removed by model scale or training approach.

\subsection{Bias Measurement Implementation Notes}
\label{sec:bias_measurement}

The measurement setup, neutral center $\mu^{(l)}$, and normalized directional projection $b_i^{(l)}$ are defined in Section~\ref{sec:spectral_energy}'s sibling subsection on representation-level position bias. We list here the numerical and edge-case details omitted from the main text.

\paragraph{Numerical constants.} We set $\epsilon_b = 10^{-8}$ inside the projection denominator for numerical stability. If $\lVert d^{(l)} \rVert$ or $\lVert v_i^{(l)} \rVert$ falls below the degeneracy threshold $\tau_b = 10^{-12}$, we set $b_i^{(l)} = 0$ for that sample-layer pair and log it as a skipped projection.

\paragraph{Interpretation.} The projection $b_i^{(l)} \in [-1, 1]$ measures cosine alignment between the deviation direction $d^{(l)}$ and the direction toward evidence $i$. A value of $1$ means $c^{(l)}$ aligns perfectly with evidence $i$, $0$ means orthogonality, and $-1$ means opposing alignment. Because cosine projections can be negative and do not sum to one, we treat $b_i^{(l)}$ as directional alignment strength rather than probability mass. Larger gaps (e.g., $b_1^{(l)} \gg b_2^{(l)}$) indicate stronger positional asymmetry.

\paragraph{Layer-wise computation.} We compute $b_i^{(l)}$ across all layers $l \in \{1, \ldots, L\}$ to track how bias evolves through depth, allowing us to distinguish bias that emerges in lower layers from bias that accumulates gradually across the network.

\subsection{Representation-Level Position Bias in GPT-OSS-20B}
\label{sec:bias_gpt20b}

To assess whether representation-level positional asymmetry is specific to the Llama architecture, we apply the same directional bias attribution analysis to GPT-OSS-20B. The full methodology is provided in Appendix~\ref{sec:bias_measurement}.
Figure~\ref{fig:bias_gpt20b} shows layer-wise directional projections for all six conflict types on GPT-OSS-20B.
The geometric bias toward earlier evidence persists from the lowest to the highest layers, closely mirroring the pattern observed in Llama-3.1-8B-Instruct. As shown in Figure~\ref{fig:bias_simple_prompt}, this cross-layer tendency is already evident in the Llama model.
This cross-model consistency suggests a shared pattern on tested models under our dataset: directional asymmetry in combined-evidence representations appears across both architectures, rather than being limited to a Llama-specific artifact.

\begin{figure*}[ht!]
  \centering
  \includegraphics[width=0.88\textwidth]{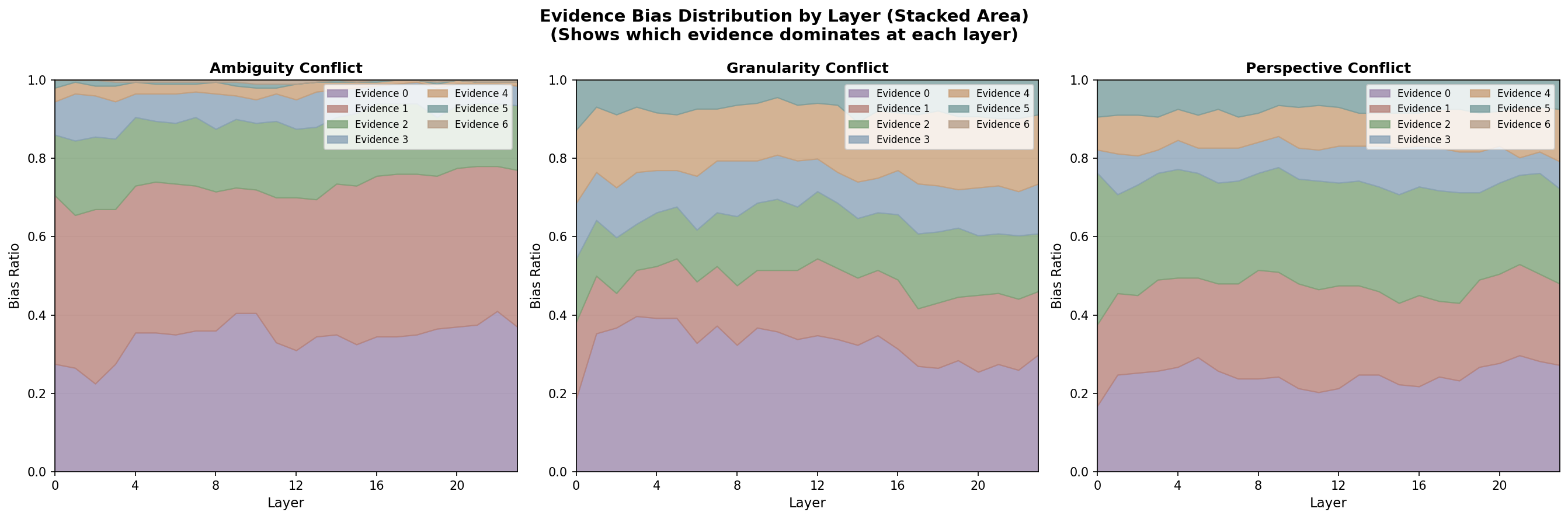}
  \caption{Layer-wise directional projection in GPT-OSS-20B via directional bias attribution. The combined representation consistently aligns with earlier evidence directions across layers and conflict types, replicating the pattern in Llama-3.1-8B and suggesting cross-model consistency of representation-level positional bias on tested models.}
  \label{fig:bias_gpt20b}
\end{figure*}

\subsection{Position Shuffling Experiment}
\label{appendix:shuffling}

To test whether randomly reordering evidence input can mitigate position bias, we permute evidence order at inference time and re-measure evidence contribution distributions using our Shapley-based attribution framework.
Tables~\ref{tab:shuffle_llama} and~\ref{tab:shuffle_gpt20b} show that Evidence~1 continues to dominate after shuffling on both Llama-3.1-8B-Instruct and GPT-OSS-20B. This result suggests that position bias is mechanistically stable and is not fully resolved by simple input-ordering heuristics in our experiments.

\begin{table}[ht]
\centering
\small
\caption{Evidence contribution distribution after position shuffling (Llama-3.1-8B-Instruct).}
\label{tab:shuffle_llama}
\resizebox{\columnwidth}{!}{%
\begin{tabular}{lcccccc}
\toprule
Conflict Type & Ev.1 & Ev.2 & Ev.3 & Ev.4 & Ev.5 & Ev.6 \\
\midrule
Ambiguity    & 59.0\% & 28.9\% & 5.9\% & 6.1\% & --- & --- \\
Granularity  & 35.3\% & 18.3\% & 14.0\% & 15.2\% & 12.6\% & 4.6\% \\
Perspective  & 47.4\% & 27.2\% & 10.9\% & 5.0\% & 3.6\% & 5.9\% \\
\bottomrule
\end{tabular}
}
\end{table}

\begin{table}[ht]
\centering
\small
\caption{Evidence contribution distribution after position shuffling (GPT-OSS-20B).}
\label{tab:shuffle_gpt20b}
\resizebox{\columnwidth}{!}{%
\begin{tabular}{lcccccc}
\toprule
Conflict Type & Ev.1 & Ev.2 & Ev.3 & Ev.4 & Ev.5 & Ev.6 \\
\midrule
Ambiguity    & 55.2\% & 32.2\% & 4.9\% & 7.5\% & --- & --- \\
Granularity  & 27.4\% & 17.4\% & 16.4\% & 16.7\% & 15.5\% & 6.5\% \\
Perspective  & 46.5\% & 22.8\% & 14.0\% & 5.7\% & 4.1\% & 6.9\% \\
\bottomrule
\end{tabular}
}
\end{table}

\subsection{Steering Strength Sensitivity Analysis}
\label{sec:alpha_ablation}

\begin{figure}[ht!]
  \centering
  \includegraphics[width=\columnwidth]{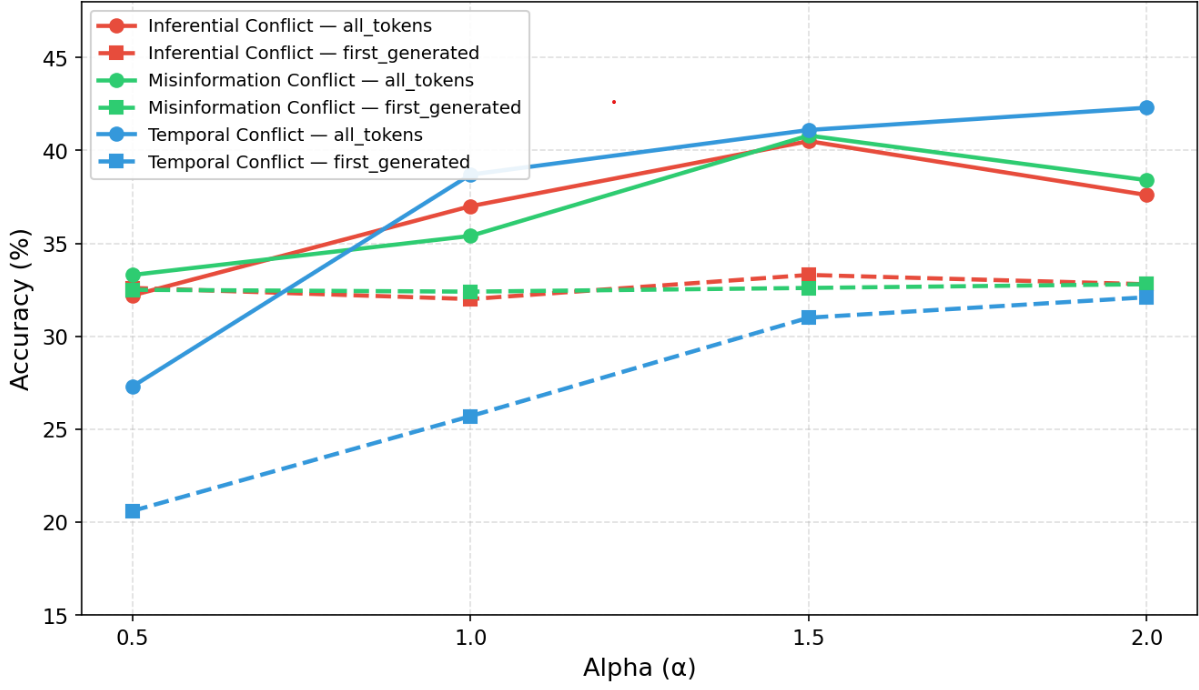}
  \caption{Sensitivity of activation steering to the coefficient $\alpha$ on Llama-3.1-8B-Instruct. Balance scores (lower is better) and Accuracy (higher is better) remain stable across a wide range of $\alpha$, suggesting robustness to this hyperparameter within the tested range.}
  \label{fig:alpha_ablation}
\end{figure}

To assess sensitivity to steering strength, we evaluate Llama-3.1-8B-Instruct across $\alpha \in \{0.5, 1.0, 1.5, 2.0\}$ on all six conflict categories.
Figure~\ref{fig:alpha_ablation} reports Balance scores (summarization tasks) and Accuracy (reasoning tasks) for each value of $\alpha$.
The curves remain stable across this range. Accuracy improves from $\alpha=0.5$ to $\alpha=1.5$ for all three reasoning conflict types under both steering variants, and changes only modestly at $\alpha=2.0$. Temporal conflicts show the largest gain, especially under \texttt{first\_generated} steering, while inferential and misinformation conflicts follow the same overall trend with smaller variation. These results indicate that the method is robust to the choice of $\alpha$, with strong performance throughout the tested range and a reliable operating region around $\alpha \in [1.0, 1.5]$.

\section{Mitigation Method Details}

\subsection{Prompts}
\label{sec:prompts}

We use different prompts for summarization and reasoning tasks. Each prompt consists of a system message and a user message. All prompts are designed to be concise and task-appropriate.

\subsubsection{Summarization Tasks}

For summarization tasks (ambiguity, granularity, perspective conflicts), we use three prompt variants shown in Figures~\ref{fig:prompt_summ_simple},~\ref{fig:prompt_summ_fairness}, and~\ref{fig:prompt_summ_single}.

\begin{figure*}[t]
  \centering
  \begin{tcolorbox}[
    enhanced,
    colback=blue!5!white,
    colframe=blue!75!black,
    fonttitle=\bfseries,
    fontupper=\small,
    title=Simple Prompt for Summarization,
    rounded corners,
    width=0.95\linewidth
  ]
  \textbf{System:} You are a helpful assistant that answers questions based on provided evidence. Answer concisely in 2--3 sentences.\\[6pt]
  \textbf{User:} \{evidence\_text\}\\
  \textbf{[Question]:} \{question\}
  \end{tcolorbox}
  \caption{Simple prompt for summarization tasks.}
  \label{fig:prompt_summ_simple}
\end{figure*}

\begin{figure*}[t]
  \centering
  \begin{tcolorbox}[
    enhanced,
    colback=blue!5!white,
    colframe=blue!75!black,
    fonttitle=\bfseries,
    fontupper=\small,
    title=Prompt for Summarization,
    rounded corners,
    width=0.95\linewidth
  ]
  \textbf{System:} You are a neutral information synthesis assistant. Analyze all sources with equal importance and provide unbiased, balanced answers.\\[6pt]
  \textbf{User:} \{evidence\_text\}\\
  \textbf{[Question]:} \{question\}
  \end{tcolorbox}
  \caption{Prompt for summarization tasks.}
  \label{fig:prompt_summ_fairness}
\end{figure*}

\begin{figure*}[t]
  \centering
  \begin{tcolorbox}[
    enhanced,
    colback=blue!5!white,
    colframe=blue!75!black,
    fonttitle=\bfseries,
    fontupper=\small,
    title=Single-Evidence Prompt for Summarization,
    rounded corners,
    width=0.95\linewidth
  ]
  \textbf{System:} You are a helpful assistant. Answer the question based ONLY on the provided evidence. Be concise and factual.\\[6pt]
  \textbf{User:} \textbf{[Evidence]:} \{source\_content\}\\[4pt]
  \textbf{[Question]:} \{question\}
  \end{tcolorbox}
  \caption{Single-evidence prompt for summarization tasks.}
  \label{fig:prompt_summ_single}
\end{figure*}

\subsubsection{Reasoning Tasks}

For reasoning tasks (inferential, misinformation, temporal conflicts), we use prompts shown in Figures~\ref{fig:prompt_reason_single} and \ref{fig:prompt_reason_combined}.

\begin{figure*}[t]
  \centering
  \begin{tcolorbox}[
    enhanced,
    colback=orange!5!white,
    colframe=orange!75!black,
    fonttitle=\bfseries,
    fontupper=\small,
    title=Single-Evidence Prompt for Reasoning,
    rounded corners,
    width=0.95\linewidth
  ]
  \textbf{System:} You are a helpful assistant that provides step-by-step reasoning.\\[6pt]
  \textbf{User:} \textbf{[Evidence]:} \{source\_content\}\\[4pt]
  \textbf{[Question]:} \{question\}\\[4pt]
  Based on the evidence above, think step by step and provide your final answer inside \texttt{<answer></answer>} tags.
  \end{tcolorbox}
  \caption{Single-evidence prompt for reasoning tasks.}
  \label{fig:prompt_reason_single}
\end{figure*}

\begin{figure*}[t]
  \centering
  \begin{tcolorbox}[
    enhanced,
    colback=orange!5!white,
    colframe=orange!75!black,
    fonttitle=\bfseries,
    fontupper=\small,
    title=Combined-Evidence Prompt for Reasoning,
    rounded corners,
    width=0.95\linewidth
  ]
  \textbf{System:} You are a helpful assistant that provides step-by-step reasoning.\\[6pt]
  \textbf{User:} \textbf{[Evidence 1]:} \{source\_1\}\\[4pt]
  \textbf{[Evidence 2]:} \{source\_2\}\\
  ...\\[4pt]
  \textbf{[Question]:} \{question\}\\[4pt]
  Analyze all evidence above and answer the question. Think step by step, then provide your final answer inside \texttt{<answer></answer>} tags.
  \end{tcolorbox}
  \caption{Combined-evidence prompt for reasoning tasks.}
  \label{fig:prompt_reason_combined}
\end{figure*}

\end{document}